\documentclass[lettersize,journal]{IEEEtran}
\usepackage{graphicx}
\usepackage{amsmath,amssymb,amsfonts}
\usepackage{bbm}
\usepackage{color}
\usepackage{multirow}
\usepackage{booktabs}
\usepackage[T1]{fontenc}
\usepackage{subfigure}
\usepackage{float}
\usepackage{setspace}
\usepackage{hyperref}
\usepackage{url}
\usepackage{makecell}
\usepackage{cleveref}
\usepackage{subcaption}
\usepackage{subfigure}
\usepackage{algorithm, algpseudocode}
\usepackage{algorithmicx}
\usepackage{setspace}
\usepackage{colortbl}
\usepackage{ragged2e}
\usepackage[table]{xcolor}
\usepackage[usenames,dvipsnames]{xcolor}

\def\BibTeX{{\rm B\kern-.05em{\sc i\kern-.025em b}\kern-.08em
    T\kern-.1667em\lower.7ex\hbox{E}\kern-.125emX}}
\usepackage{balance}

\begin{document}
\title{Benchmarking Semantic Segmentation Models via Appearance and Geometry Attribute Editing}

\author{Zijin Yin, Bing Li, Kongming Liang, Hao Sun, Zhongjiang He, \\ Zhanyu Ma,~\IEEEmembership{Senior Member, ~IEEE}, and Jun Guo, ~\IEEEmembership{Senior Member,~IEEE}

\vspace{-1em}

\IEEEcompsocitemizethanks{
    \IEEEcompsocthanksitem Corresponding author: Kongming Liang.
    \IEEEcompsocthanksitem Z. Yin, K. Liang, Z. He, Z. Ma and J. Guo are with the School of Artificial Intelligence, Beijing University of Posts and Telecommunications, and Beijing Key Laboratory of Multimodal Data Intelligent Perception and Governance, Beijing 100876, China. Z. He is also with the China Telecom Artificial Intelligence Technology Co. Ltd, Beijing 100034, China. E-mail: (yinzijin2017, liangkongming, hezhongjiang, mazhanyu, guojun)@bupt.edu.cn
    \IEEEcompsocthanksitem B. Li is with the Visual Computing Center, King Abdullah University of Science and Technology, Jeddah, Saudi Arabia. E-mail: bing.li@kaust.edu.sa
    \IEEEcompsocthanksitem{H. Sun is with the China Telecom Artificial Intelligence Technology Co. Ltd, Beijing 100034, China. E-mail: sunh10@chinatelecom.cn}
}}

\markboth{Transactions on Pattern Analysis and Machine Intelligence}%
{Shell \MakeLowercase{\textit{et al.}}: Bare Demo of IEEEtran.cls for Computer Society Journals}

\maketitle

\begin{abstract}
Semantic segmentation takes a pivotal role in various applications such as autonomous driving and medical image analysis. When deploying segmentation models in practice, it is critical to test their behaviors in varied and complex scenes in advance. In this paper, we construct an automatic data generation pipeline \textit{Gen4Seg} to stress-test semantic segmentation models by generating various challenging samples with different attribute changes. 
Beyond previous evaluation paradigms focusing solely on global weather and style transfer, we investigate variations in both appearance and geometry attributes at the object and image level. These include object color, material, size, and position, as well as image-level variations such as weather and style.
To achieve this, we propose to edit visual attributes of existing real images with precise control of structural information, empowered by diffusion models. In this way, the existing segmentation labels can be reused for the edited images, which greatly reduces the labor costs of constructing datasets. 
Using our pipeline, we construct two new benchmarks, Pascal-EA and COCO-EA. We benchmark a broad variety of semantic segmentation models, spanning from conventional close-set models to recent open-vocabulary large models. We have several key findings: 
1) advanced open-vocabulary models do not exhibit greater robustness compared to closed-set methods under geometric variations;
2) traditional data augmentation techniques, such as CutOut and CutMix, are limited in enhancing robustness against appearance variations;
3) our generation pipeline can also be employed as a data augmentation tool and improve both in-distribution and out-of-distribution performances.
Our work suggests the potential of generative models as effective tools for automatically analyzing segmentation models, and we hope our findings will assist practitioners and researchers in developing more robust and reliable segmentation models. Codes and datasets are available at \url{https://github.com/PRIS-CV/Pascal-EA}.
\end{abstract}


\section{Introduction}
\label{sec:introduction}
\IEEEPARstart{S}{emantic} segmentation is a fundamental task and plays a pivotal role in a wide range of applications. In the last few years, deep-learning-based semantic segmentation models have witnessed remarkable progress in addressing complex visual scenes, enabling ubiquitous applications ranging from autonomous driving \cite{autonomous-driving} to medical image analysis \cite{medicalsam}. In order to ensure their reliability and robustness in real-world scenarios, it is desirable to evaluate them in varied and complex scenes in advance.

\begin{figure}
    \centering
    \includegraphics[width=\columnwidth]{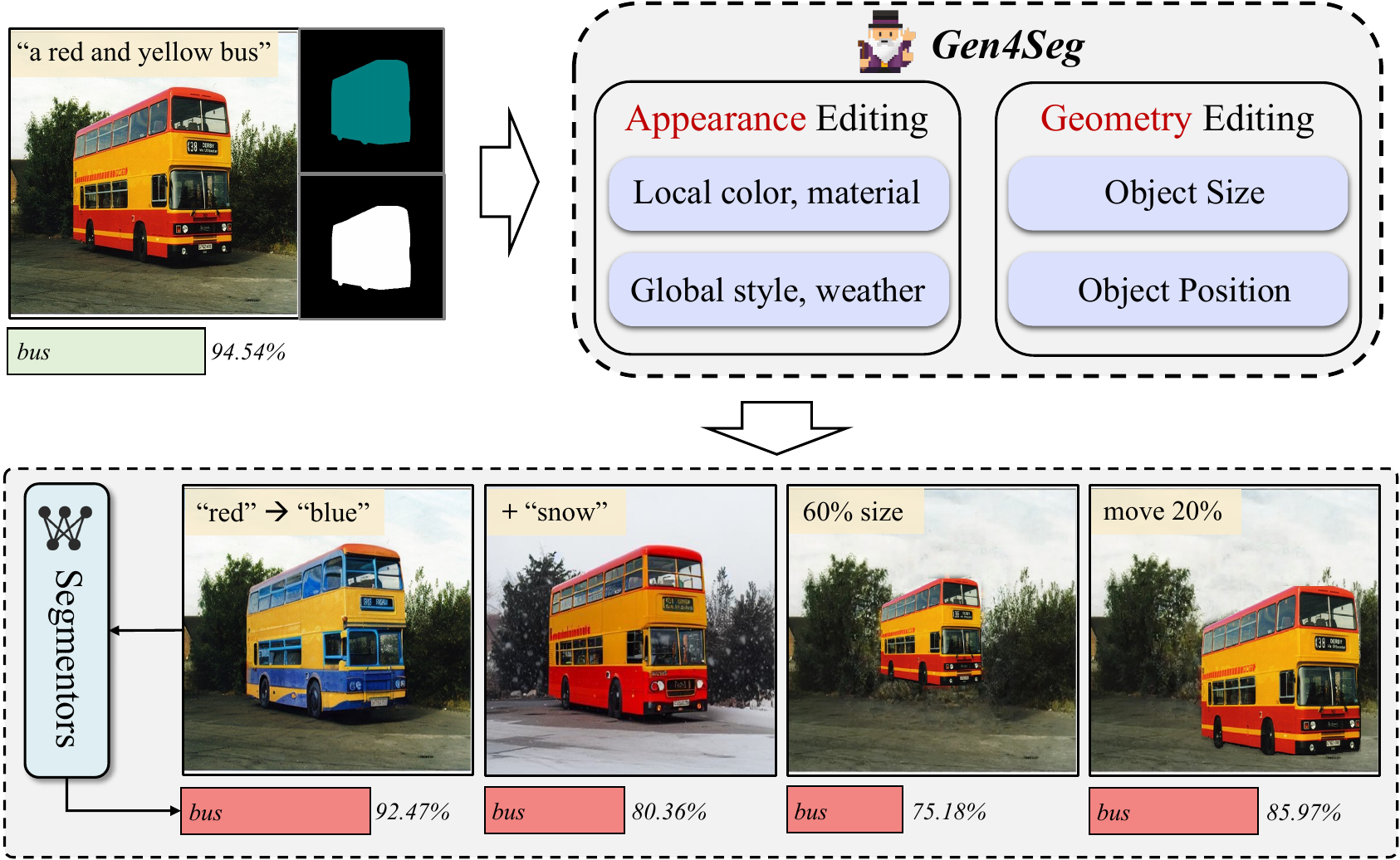}
    \caption{\textit{Gen4Seg} expands existing images with various appearance and geometry attributes changes for evaluating the robustness of semantic segmentation models.}
    \vspace{-1.6em}
    \label{fig:teaser}
\end{figure}

As to varied real-world scenarios, segmentation methods are required to be robust to various shifts in terms of appearance attribute variations (\textit{e.g.} object color, material, weather conditions) and geometry attribute changes (\textit{e.g.} object size and position). For example, a bus can be composed of different colors, or placed in different weather conditions (see Fig. \ref{fig:teaser}). Moreover, the geometric layout of objects and scenes differs vastly in the real world. Nevertheless, the problem is: \textit{``How sensitive are existing segmentation models to appearance and geometry attribute changes?"} Regrettably, this research avenue remains under-explored.

The main challenge of robustness evaluation is the lack of high-quality test data with abundant diversity. In addition, even if the data can be collected, the annotation cost of the segmentation masks is quite high. 
The ACDC dataset \cite{acdc} manually collects samples with adverse weather in city streets. Sun \textit{et al.} \cite{shift} used a simulation software to build a synthetic dataset, while Multi-weather \cite{multiweather} and Stylized-COCO \cite{trapped} employed style-transfer models \cite{cyclegan} to generate data. However, most of them only evaluate segmentation models under global variations, such as weather and different road scenes in autonomous driving scenarios. Moreover, due to the limited capabilities of the simulation software itself and conventional style-transfer models, their data are not consistent with realistic images and hence impede the provision of reliable assessment results.

Recently, capitalizing on the remarkable capability of diffusion models \cite{ldm} in content generative, researchers have devoted their efforts \cite{lance,imagenetd} to evaluating the robustness and generalization ability of image recognition models. They use existing labels to augment test samples across more general scenes. For example, ImageNet-E \cite{li2023imagenet} introduces a variant of ImageNet \cite{imagenet} by modifying object size, position, rotation, and background content using diffusion models \cite{ddim, ddpm}. Prabhu \textit{et al.} \cite{lance} leverage state-of-the-art text-based image editing techniques, Prompt-to-Prompt \cite{p2p}, to augment real test images with various content changes, stress-testing classification models. ImageNet-D \cite{li2023imagenet} synthesizes diverse test images by modifying backgrounds, textures, and material appearances through diffusion model prompt augmentation \cite{ldm}. However, their works are limited to classification tasks.

In this paper, we investigate the robustness of semantic segmentation models under varying visual attribute conditions. To achieve this, we provide a data generation pipeline, named \textit{Gen4Seg}, which produces synthetic data encompassing diverse variations, including both appearance (color, material, style, and weather) and geometry (size and position) attributes (see Fig. \ref{fig:teaser}). 
Drawing inspiration from the impressive capabilities of text-guided image editing methods \cite{pnp, p2p, masactrl}, we employ a pre-trained diffusion model with text instructions to transform test images by editing their attributes. To avoid the need for manual annotation of labels for these edited images, it is crucial to preserve the structural integrity during editing, such that the ground-truth segmentation mask of an original image is allowed to use to its edited ones. 
However, typical methods often improperly modify irrelevant contents (\textit{e.g.} removing an object from the background, changing the target object shape) during editing. 
To address this issue, we propose a mask-guided energy function in the diffusion model \cite{ldm} to consistently edit target appearance attributes at the object level and preserve the structure of an image. Additionally, we incorporate ControlNet \cite{controlnet} and Visual Language Models (VLMs) to enable faithful and automated editing. To further reduce noise and artifacts in the generated images, we introduce a two-stage automatic noise filtering strategy. Thanks to these modules, our pipeline performs faithful attributes editing in a tuning-free manner, while avoiding manually annotating ground-truth segmentation masks.

Based on our \textit{Gen4Seg}, we construct two new benchmarks Pascal-EA (\textbf{E}ditatble \textbf{A}ttributes) and COCO-EA covering various object and image attribute changes. By extensive experiments, we have three key findings: 
1) Advanced open-vocabulary models, having stronger backbones and being trained on massive datasets, do not exhibit higher robustness than closed-set methods under geometry attribute variations. 
2) Traditional data augmentation methods show limited effectiveness when handling appearance attribute variations. 
3) Utilizing our pipeline as a data augmentation tool also enhances robustness in both in-distribution and out-of-distribution scenarios. 

The main contributions are summarized as follows:
\begin{itemize}
    \item We provide an automatic data generation pipeline \textit{Gen4Seg}, capable of editing real-image appearance and geometry attributes in a tuning-free manner without the requirement of re-collecting segmentation labels.
    \item We introduce 2 new benchmarks, Pascal-EA and COCO-EA, each owns 8 groups that cover object, image appearance, and object geometry variations.
    \item We comprehensively analyze 13 semantic segmentation models and 4 popular data augmentation methods. We summarize several effective findings, and it is identified that object appearances and geometry variations should be carefully considered in future research and real-world deployment. 
\end{itemize}

A preliminary version of this work was published in \cite{yin2024benchmarking}. Our new contributions include: (1) an additional object geometry editing method for modifying object size and position attributes (Sec. \ref{subsec:geo_edit}); (2) an updated strategy for local object appearance editing (Sec. \ref{subsec:appear_edit}); (3) new benchmarks with an added dataset and two additional variation types (Sec. \ref{sec:dataset}); (4) more analysis to assess the reliability of our pipeline, and (5) more benchmark results and findings.
\begin{figure*}[t]
    \centering
    \includegraphics[width=\linewidth]{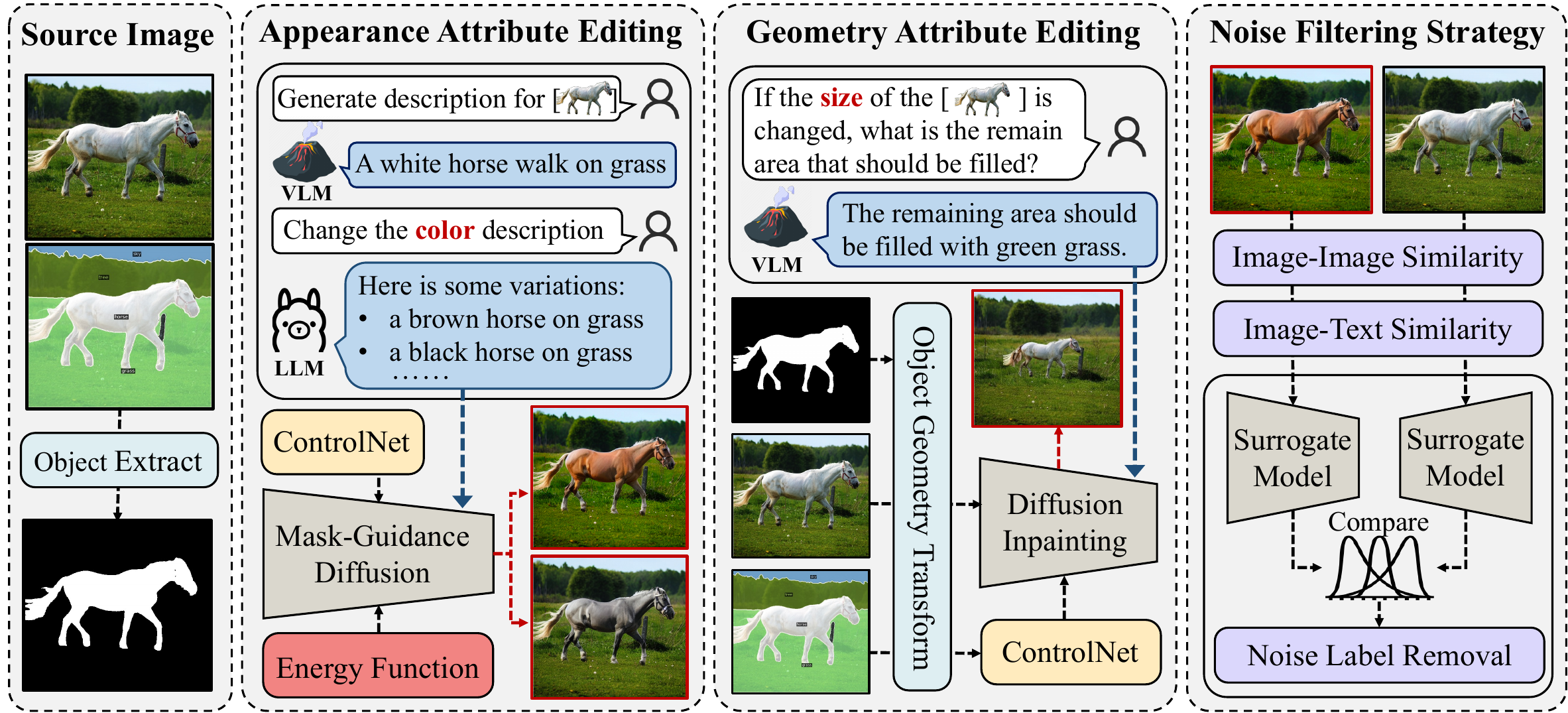}
    \caption{\textbf{Our \textit{Gen4Seg} generates appearance and geometry attributes variations to existing images, with collaboration of diffusion models, VLMs and LLMs.} For appearance editing, we employ a VLM to edit text description of target object, and use proposed Mask-Guidance Diffusion model with ControlNet and designed energy function. For geometry editing, we first manipulate (move, rescale) the intent edit object, and use a VLM and ControlNet to recover background content by a diffusion-based inpainting model. At last, we apply established metrics to discard low-quality images and further employ a surrogate segmentation model to eliminate harmful regions from evaluation.}
    \vspace{-1.5em}
    \label{fig:pipeline}
\end{figure*}

\section{Related Work}
\noindent \textbf{Robustness evaluation in Semantic Segmentation.} 
Several efforts have been made to evaluate the robustness of segmentation models under domain shifts caused by specific attribute changes \cite{acdc,shift,multiweather,foggydriving,robust,loiseau2024reliability}. The ACDC benchmark \cite{acdc} manually collects images and labels from multiple adverse weather conditions. SHIFT \cite{shift} employs simulation software to automatically render data with continuous and discrete scene shifts in city streets. Other works \cite{adain, multiweather, foggydriving, robust} synthesize images by training style transfer models \cite{cyclegan, stylegan}. GenVal \cite{loiseau2024reliability} also uses diffusion models with layout controls to generate diverse weather conditions and Out-of-Distribution objects in urban street scenes. However, these approaches primarily focus on global attribute variations, such as weather \cite{acdc} or scene changes \cite{shift} in autonomous driving contexts.
Theodoridis \textit{et. al.} \cite{trapped} explores texture bias by altering object or background styles through a style transfer model, but it can only modifies local textures, and generated images are highly unrealistic. In contrast, our research incorporates both local and global attribute variations, focusing on real-world scenarios, which allows us to more comprehensive evaluations.

\noindent \textbf{Model Diagnosis with Synthetic Data.} 
Diagnosing the model's behavior and explaining its mistakes is critical for understanding and building robustness. 
Counterfactual explanation \cite{counterfactualreview,tabular} is to explain deep vision models' behavior \cite{counterfactual,steex, beyond} by establishing the human-interpretable link between model predictions and perturbations to input images.
Recently, generative approaches \cite{texture-bias, discover-bugs, diffusion-explanation, khorram2022cycle, luo2023zero, zemni2023octet, li2023imagenet, lance, vendrow2023dataset, jeanneret2023text, howard2023probing, imagenetd} achieve remarkable progress in explaining model failures. 
Luo \textit{et al.} \cite{luo2023zero} measure models' sensitivity to pre-defined attribute changes by optimization in latent space of StyleGAN \cite{stylegan}.
ImageNet-E \cite{li2023imagenet} changes object size, position, and rotation by diffusion models \cite{ddim,ddpm} to evaluate the robustness of classification models.
Prabhu \textit{et al.} \cite{lance} and ImageNet-D \cite{imagenetd} synthesize test images with diverse appearances by diffusion models \cite{ldm}, and evaluate large recognition model CLIP \cite{clip} and MiniGPT-4 \cite{minigpt4}.
Chen \textit{et al.} \cite{chen2025model} also employ editing techniques to produce counterfactuals and use them to diagnose and correct models' bias.
Yi \textit{et al.} \cite{yi2024benchmarking} evaluate the robustness of optical flow estimation methods under various perturbations, such as noise and motion blur. 
However, these approaches are unsuitable for image segmentation tasks, as their editing methods interfere with irrelevant regions or require pixel labels to be recollected after generation.
In contrast, our proposed method leverages existing labels to avoid such interference and generate more realistic and faithful test images without recollecting labels.

\noindent \textbf{Image Editing.} 
Diffusion models offer an efficient way to synthesize and edit images conditioned on text prompts. \cite{p2p, null, pnp, masactrl,xu2023inversion,lin2024text} edit both global and local aspects of the image by injecting the feature extracted from real images and controlling the attention maps in the forward diffusion process. However, their method inevitably interferes with other regions in local editing. 
For geometry editing, ImageNet-E \cite{imagenet} utilizes an unconditional latent diffusion model; thus, it suffers from recovering background content. Recently, some works \cite{shi2024dragdiffusion,yu2025objectmover}, such as DragDiffusion \cite{shi2024dragdiffusion}, can move objects, but they can not accurately modify object positions, thus can not be applied to our work. These issues are fatal in our setting, since segmentation methods may collapse under any unnecessary pixel perturbations, and generated variations need to strictly correlate to the original images. Our proposed approach uses extra masks to prevent potential interference and generate more faithful image variations.

\vspace{-0.3em}
\section{Method}
\label{sec:method}
In this section, we present a novel generative pipeline to synthesize appearance and geometry attributes variations, using the collaboration of Vision-Language Models (VLMs), Large Language Models (LLMs), and Generative Diffusion Models. The overall pipeline is illustrated in Fig. \ref{fig:pipeline}. We first introduce preliminaries on diffusion model. Next, we describe the entire pipeline, which includes the proposed appearance and geometry attribute editing approaches. Finally, we introduce our two-stage noise filtering strategy.

\subsection{Preliminaries}
\noindent \textbf{Diffusion model.}
Inspired by non-equilibrium thermodynamics \cite{nonequilibrium}, diffusion models \cite{ddim, ddpm, improvedddpm} are probabilistic generative models that slowly add random noise to data and then learn to reverse the diffusion process to construct desired data samples from the noise. 
Formally, given an original data sample $z_{0}$, the forward diffusion process successively generates noised variants $z_1, z_2 \dots, z_T$ by a Markov-Chain: 
\begin{equation}
    z_t = \sqrt{\alpha_t} \cdot z_{t-1} + (1-\alpha_t) \cdot \epsilon_{t-1}
\end{equation}
where $\epsilon_{t-1} \sim \mathcal{N}(0, \mathbf{I})$, $t \in\{1, \ldots, T\}$ and $T$ is the number of diffusion steps, $\alpha_t \in[0,1)$ is predefined noise adding schedule. The data sample gradually loses its distinguishable features as the step $t$ increases. Eventually, as $T \to \infty$, $z_T$ is equivalent to an isotropic Gaussian distribution. 
The reverse diffusion process uses a neural network $\epsilon_\theta(z_t, t)$ to predict noise $\epsilon$ in each step $t$, and generates a clean sample $z_{t-1}$ from $z_t$ by removing the predicted noise.



\noindent \textbf{DDIM Inversion.} 
The goal of inversion is to reverse the generation process of a pre-trained diffusion model.
Given a real image, the DDIM inversion \cite{ddim, null} process approximates the corresponding noise that would reconstruct the same image during the diffusion forward process. Based on the assumption that Ordinary Differential Equation (ODE) process can be reversed in the limit small steps, DDIM inversion performs as:
\begin{equation}
\begin{split}
    z_{t} &= \sqrt{\frac{\alpha_{t}}{\alpha_{t-1}}} z_{t-1} \\
    &+ \sqrt{\alpha_{t}} \left( \sqrt{\frac{1}{\alpha_{t}} - 1} 
    - \sqrt{\frac{1}{\alpha_{t-1}} - 1} \right) \cdot \epsilon_\theta (z_{t-1},t)
\end{split}
\end{equation}
It allows backward traversal from real images to deterministic noises and are widely used in image editing.

\noindent \textbf{Prompt-to-Prompt (P2P) \cite{p2p}} introduces an innovative image editing techniques. It reveals coarse localization ability to semantic objects within cross-attention maps between text token embeddings and image features. By adjusting the values of these cross-attention maps or injecting new ones, it enables fine-grained control over images solely through textual modifications. 

\subsection{Appearance Attribute Editing}
\label{subsec:appear_edit}
We begin by categorizing common appearance attributes observed in the real world. Next, we modify text descriptions to reflect changes in these appearance attributes and edit the images using our proposed Mask-Guided Diffusion model. Through this approach, we can synthesize images featuring diverse objects and styles in an end-to-end manner.

\subsubsection{Appearance Attribute Taxonomy} 
Generally, appearance attributes \cite{farhadi2009describing, liang-2019pami, liang2023hierarchical} refer to the characteristic properties of objects and images, including color, shape, texture, and style, among others. A visual instance can be described by a combination of various attributes. For example, a cat might exhibit attributes such as \textit{brown color, soft texture, dotted skin, and feline shape}. In existing visual datasets, certain attributes, like the shape of a cat, remain relatively constant across diverse environments, while others, such as texture, material, and color, display greater variation within a specific class \cite{net2vec}. To generate diverse evaluation samples, we focus on attributes that typically vary in real-world scenarios, ranging from local objects to entire scenes.
Specifically, we manually select attributes from four categories: (a) Color, \textit{e.g.}, blue, red. (b) Material, \textit{e.g.}, metal, wood, stone. (c) Style, \textit{e.g.}, photo, painting, sketch. (d) Weather conditions, \textit{e.g.}, snow, rain, fog. By modifying these appearance attributes, we can generate abundant test samples with variations.

\subsubsection{Text Editing} 
Since existing diffusion generative models are usually instructed by language, we first obtain the textual description of a real test image, and use a LLM to edit specific descriptions of different attributes. In this way, our pipeline can generate synthetic images in an end-to-end manner without human involvement. 
Specifically, for a given real image $I$, we first use a pretrained Vision-Language Model (LLaVA \cite{llava, llava2}) to acquire its text description $P$. 
To perform meaningful modification to text, we divided a text description into several editable components based on linguistic formal: domain, subjects and their adjectives (attributes), actions (verbs), and backgrounds (objects). For example, in \textit{``a photo of a white horse on the grass''}, the domain is \textit{``a photo''}, the subject is \textit{``horse''} and its adjective is \textit{``white''}, the action is \textit{``on''} and the background is \textit{``grass''}. 
Then we employ the famous LLaMA3 \cite{llama, llama3} to generate text variation $P^*$, where there is only difference in specific attribute descriptions, by predefined instruction prompts. 
Instruction prompts for different categories attributes are shown in the supplementary material. Then, we use $P^*$ as the input prompt to guide image editing.


\begin{figure}
    \centering
    \includegraphics[width=\columnwidth]{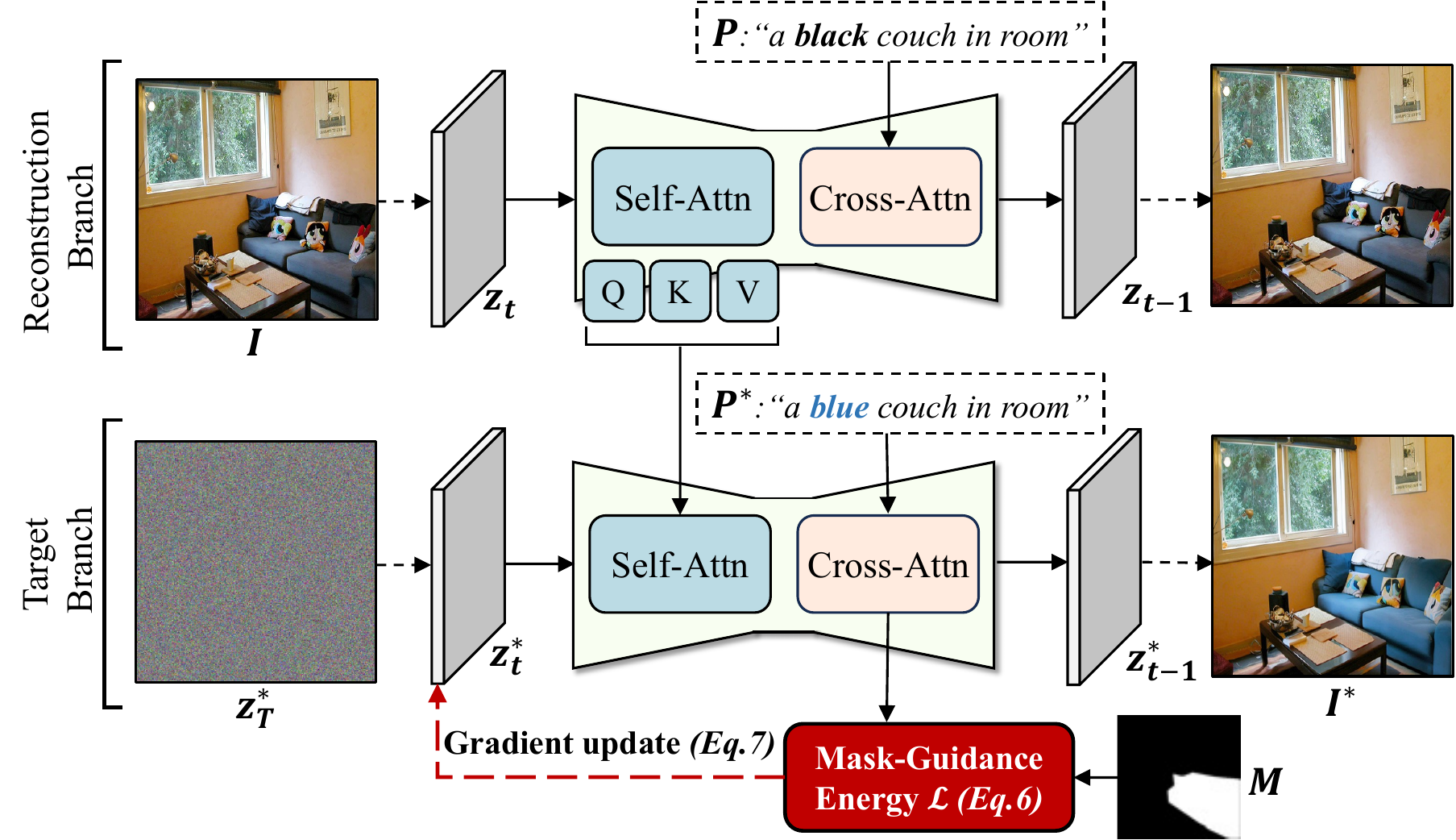}
    \caption{Illustration of our Mask-Guidance Diffusion. We propose the Mask-Guidance Energy function to optimize cross-attention maps region of target edit token.}
    \vspace{-2em}
    \label{fig:energy_function}
\end{figure}

\subsubsection{Image Editing with Mask-Guidance Diffusion}
\label{subsubsec:mask_diffusion}
\noindent \textbf{Motivation.} 
Since the generated images are used as test samples, they should have the following properties: (1) Correctness: generated images should not contravene spatial layout defined by original semantic segmentation labels. (2) Fidelity: generated images should not disrupt irrelevant information and adhering to real-world conditions.

However, current mask-free image editing approaches \cite{p2p, pnp, masactrl, zero, muse} can modifies objects and images appearances attributes, but they struggle to precisely align the local editing region with the indented texts under complex scenarios since they solely rely on the text's obscure localization capabilities \cite{p2p} in cross-attention maps. Thus they inevitably interfere irrelevant areas or generate artifacts on object boundaries, \textit{e.g.} editing object color will marginally change surrounding details. 
Please refer to Fig. \ref{fig:edits_compare} for more examples. We call this phenomenon ``edits leakage''.
The edits leakage heavily diminishes the assessment reliability of generated test images, as the segmentation performance on whole images is susceptible to any pixel perturbation \cite{segattack}. \cite{yin2024benchmarking} restricts edit region aligned with the textual prompt by multiplying pixel mask with cross-attention maps, but they can only constrain attention maps to local region but cannot rectify the deviation of attention maps. Therefore, we propose a new mask-guidance strategies for image editing. 

\noindent \textbf{Overview.} 
Suppose that the real images is $I \in \mathbb{R}^{3 \times H \times W}$ with height $H$ and width $W$, then we first compute its initial noise $z_T^G$ by DDIM inversion \cite{ddim}. 
For textual description $P$ and the target text prompt $P^*$, $S^* = \{s_1^*, s_2^*, s_n^*, ..., s_N^*\}$ is a set of tokens in $P^*$, $S' = \{s'_1, s'_2, s'_j, ..., s'_J\}$ is a set of intent editing tokens and $S' \in S^*$. 
In addition, a pixel mask $M$ is provided to accurately specify the editing region in our target image intended $I^*$. 
We adopt the framework of PnP \cite{pnp}, which consists of two branches, \textit{i.e.} a reconstruction branch starting from $z_T^G$ to  $z_0$, and an editing branch with the same initial noise $z_T^* = z_T^G$. 
Our proposed method aims to achieve better consistency between aligned editing region of $S^*$ with desired region specified by $M$. To achieve this, we first inject self-attention maps and features. Then we employ our proposed mask-guidance energy function for optimizing noise latent features, and finally employ ControlNet \cite{controlnet} to ensure semantic segmentation label correctness after editing. The overall architecture of our proposed Mask-Guidance Diffusion is illustrated in Fig. \ref{fig:energy_function}. The algorithm details are shown in the supplementary material.

\noindent \textbf{Feature and Attention Injection.} 
To preserve the spatial layouts of original image, we first employ the injection mechanism similar to that used in PnP \cite{pnp}. Specifically, intermediate features in reconstruction branch are injected into the editing branch at each step $t$ as follows:
\begin{equation}
    f^*_t = 
    \begin{cases} 
        f_t & \text{if } t \leq \theta_f \\
        f^*_t & \text{otherwise}
    \end{cases}
\end{equation}
where $f_t$ and $f^*_t$ are intermediate features of the reconstruction branch and editing branch at step $t$, respectively. Similarly, self-attention maps injection is performed as:
\begin{equation}
    \widetilde{A}^*_t = 
    \begin{cases}
        \widetilde{A}_t & \text{if } t \leq \theta_a \\
        \widetilde{A}^*_t & \text{otherwise}
    \end{cases}
\end{equation}
where $\widetilde{A}_t$ and $\widetilde{A}^*_tt$ represent the self-attention maps from the reconstruction branch and the editing branch at step $t$, respectively. $\theta_f$ and $\theta_a$ are time step thresholds that determine when $f_t$ and $A_t$ are injected. Finally, the denoising network predicts the noise using the overridden features and self-attention maps as $\epsilon_\theta(z_t, t, \{f^*_t, \widetilde{A}^*_t\})$. 

\noindent \textbf{Mask-Guidance Energy Function.} 
Score-based diffusion model \cite{song2020score} utilizes gradient guidance generated by a designed energy function that matches the editing target, and additional gradient guidance will change the diffusion path to reach the editing target. 
Inspired by the previous effort \cite{chen2024training} on layout-control, our energy function are designed to provide gradient guidance for rectifying the region of cross-attention maps to a precise localization mask. Thus, the mask-guidance energy function can be formulated as follows:
\begin{equation}
\label{eq:energy_function}
    \mathcal{L} = \left( 1 - \frac{1}{\sum{M}} \sum_{j=1}^{J} M \odot \frac{(\hat{A}_t^*)_j}{\sum_{n=1}^{N}(\hat{A}^*_t)_n} \right)^2
\end{equation}
where $(\hat{A}_t^*)_j$ represents the cross-attention map between the intent edit token $s'_j$ and image features, $(\hat{A}^*_t)_n$ represents the cross-attention map of the token $s^*_n$. And $\sum{M}$ is the summary number of elements within the mask.

Optimizing the energy function $\mathcal{L}$ encourages higher values of cross-attention maps of intent edit tokens inside the area specified by the mask $M$, and lower values outside. Then the gradient of $\mathcal{L}$ is computed to update the latent $z^*_t$ as follows:
\begin{equation}
    z^*_t \leftarrow z^*_t - \eta \nabla_{z^*_t} \mathcal{L}
\end{equation}
where $\eta$ is a scale factor controlling the strength of the gradient guidance.

Finally, we replace latent values outside the mask $M$ with the corresponding source latent at each denoising time-step $t$:
\begin{equation}
\label{eq:blend}
    z^*_t \leftarrow M \odot z^*_t + (1 - M) \odot z_t
\end{equation}
This blend operation enables the recovery of the original content in the unplanned edit region. We apply the loss on the cross-attention maps extracted from the first block of the up-sampling stage in the U-Net \cite{unet}, which we have found to be the optimal setting for balancing controllability and image quality. Additionally, this loss is applied exclusively in local object appearance editing.

\noindent \textbf{ControlNet.}
Let $G$ be the pixel-level semantic segmentation label of $I$. In order to further control that the generated image $I^*$ rigorously adheres to $G$, we integrate the existing controllable generative module ControlNet \cite{controlnet} into each block. 
In this way, our pipeline can obtain samples with abundant diversity, without changing their original semantic segmentation labels. And we can hence expand existing benchmarks requiring no laborious label collection.

\subsection{Geometry Attribute Editing}
\label{subsec:geo_edit}

\begin{figure}[t]
    \centering
    \includegraphics[width=\columnwidth]{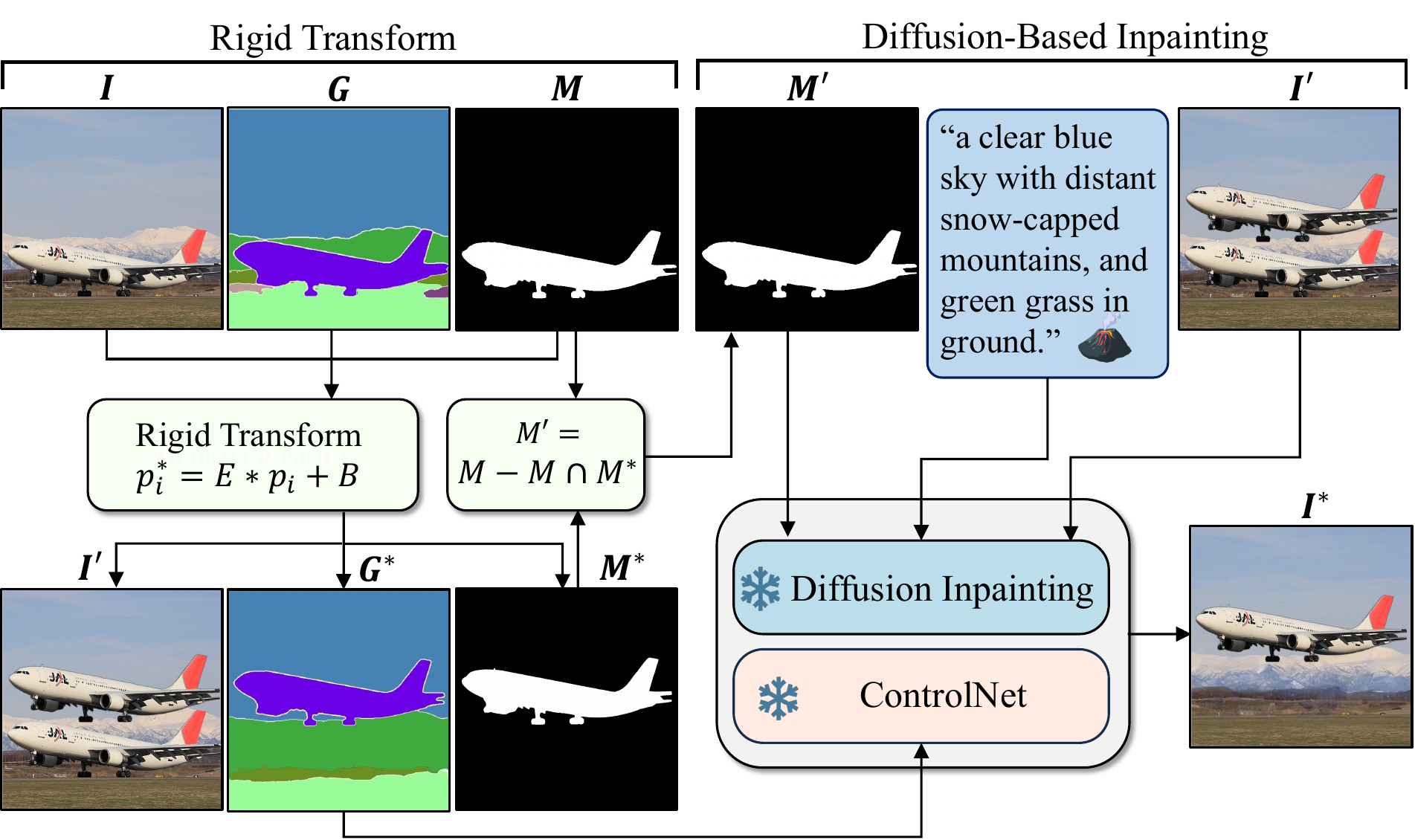}
    \caption{Illustration of our Geometry Attributes Editing pipeline. We first apply a rigid transform, then employ a diffusion inpainting model and ControlNet \cite{controlnet}.}
    \vspace{-1.5em}
    \label{fig:geo_edit}
\end{figure}

\begin{figure*}[t]
    \centering
    \includegraphics[width=\textwidth]{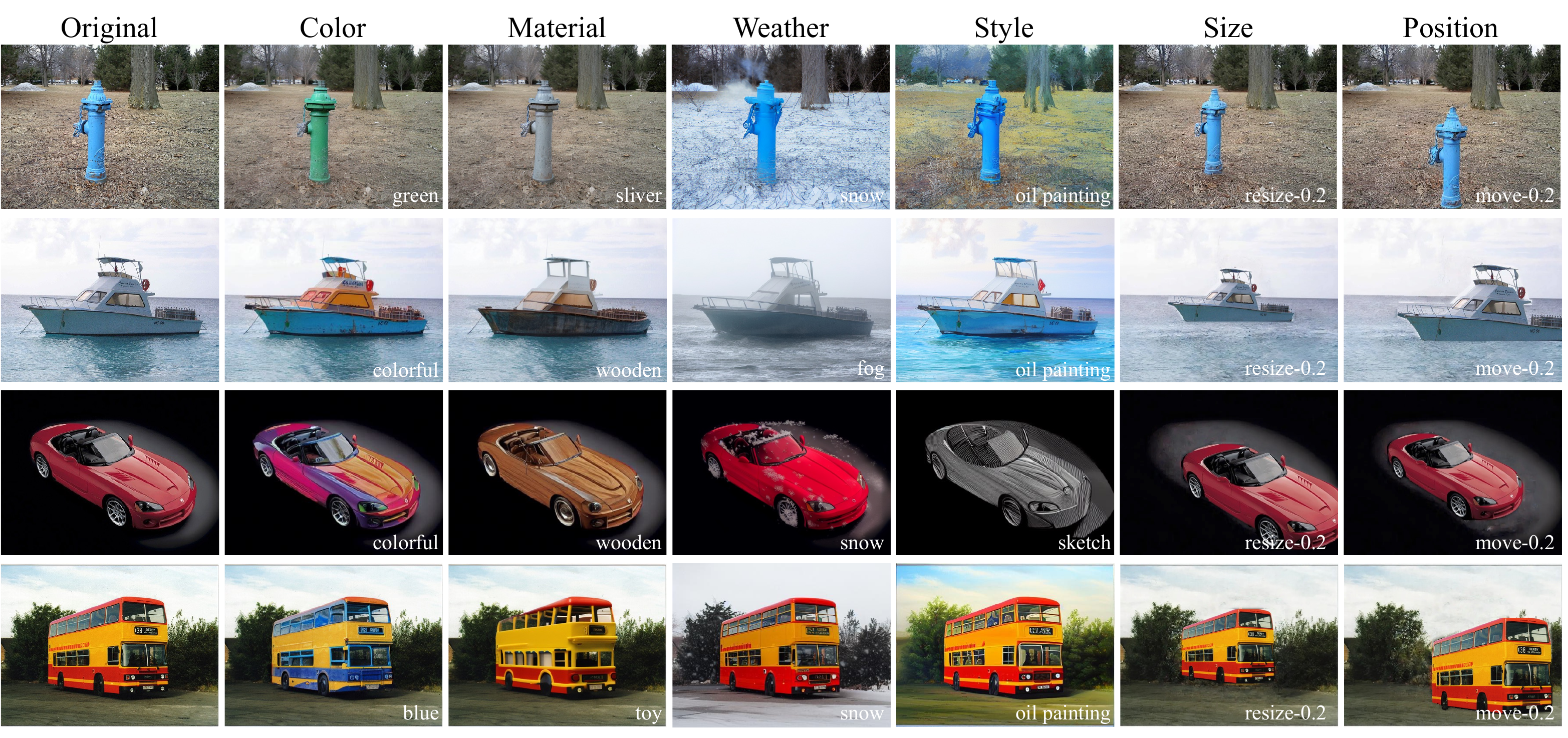}
    \caption{Examples of our generated datasets Pascal-EA and COCO-EA with diverse appearance and geometry variations.}
    \vspace{-1em}
    \label{fig:dataset_vis}
\end{figure*}

\noindent \textbf{Motivation.} 
The variations in geometry attributes at object levels impact the human perception to accurately separate scenes, we argue that this phenomenon is also established in semantic segmentation models. To evaluate models' sensitivity to geometry variations, we first define two factors, \textit{i.e.} object scale and object position, to describe geometry attributes of objects within images.
ObjectNet \cite{objectnet} examined the robustness of object recognition by controlling object geometries. However, their dataset falls outside the original distribution, making it difficult to determine whether performance degradation stems from changes in data distribution or object geometries. 
ImageNet-E \cite{li2023imagenet} introduced a diffusion-based editing tool to generate in-distribution data with variations in object size, position, and rotation. However, this approach inevitably interferes with irrelevant regions and is hence unsuitable for image segmentation tasks. Therefore, we propose a new object attribute tool for rescaling and repositioning intent editing objects. 

\noindent \textbf{Overview.} 
In source image $I$, $p_i = \begin{bmatrix} x_i & y_i \end{bmatrix}^\top$ is coordinate of $i$-th pixel within the intent edit object specified by mask $M$. We first apply a rigid transform to each $p_i$. For repainting contents in the original region of the object, we employ a diffusion-based inpainting model \cite{ldm} with ControlNet \cite{controlnet} and VLM \cite{llava, llava2} text guidance to produce coherent and faithful results. Fig. \ref{fig:geo_edit} shows the overall procedure.

\noindent \textbf{Rigid Transform.} 
We denote the rigid transformation as $p'_i = E \cdot p_i + B$ where $p'_i$ is the new coordinate of the $i$-th pixel. The matrix $E$ controls the object scale, and $B$ controls the object position.
\begin{equation}
    E = \begin{bmatrix} e_x & 0 \\ 0 & e_y \end{bmatrix}, B = \begin{bmatrix} b_x \\ b_y\end{bmatrix}
\end{equation}
where $e_x$ and $e_y$ are random scaling factors in the $x$ and $y$ directions, and $b_x$ and $b_y$ are translation factors along the $x$ and $y$ directions. Then we can obtain the transformed image $I'$. Similarly, we apply the transform to object mask $M$ and segmentation label $G$ to obtain the final $M^*$ and $G^*$.

\noindent \textbf{Image Repainting with VLM and ControlNet Guidance.} 
In the transformed image $I'$, the original area of the object needs to be repainted with background content. To achieve this, we employ the Stable Diffusion inpainting model \cite{ldm} to generate the final target image $I^*$. To prevent painting artifacts and implausible content, we also propose to use VLM \cite{llava,llava2} and ControlNet \cite{controlnet}.
Specifically, we ask the LLaVA \cite{llava,llava2} with prompt ``If the size or position of the [object name] is changed, what is the remaining area that should be filled?'', then acquire the response text $P'$. We obtain the mask $M' = M - (M \cap M^*)$, which represents the remaining area after modification. Then the repainting process is guided by the mask $M'$, the ControlNet \cite{controlnet} inputted by the new label $G*$, and text prompt $P'$. The usage of ControlNet \cite{controlnet}, and VLM \cite{llava,llava2} ensures that the inpainted area aligns seamlessly with the background content instead of the foreground object. 
We observe that using a dilated soft mask during the inpainting process enhances the coherence between the foreground object and the background and avoids copy-and-paste-like artifacts. To this end, we apply a dilation operation with $5 \times 5$ window and an average smoothing on the boundary of $M'$ with a $3 \times 3$ window. 
Finally, source image $I$ with its label $G$ and target image $I^*$ with $G*$ form pairs used to evaluate segmentation robustness. 
Unlike ImageNet-E \cite{li2023imagenet}, which disentangles the object from the background by diffusion inpainting and directly pastes the transformed object, our approach requires a smaller inpainting region.

\subsection{Noise Filtering Strategy}
\label{subsec:noise_filter}
\noindent \textbf{Motivation.} 
While editing image appearance and geometric attributes using diffusion models shows promise, it often introduces noisy areas and artifacts, particularly along object boundaries. These issues arise from the diffusion process, suboptimal hyper-parameter settings, and the inherent difficulty of editing certain objects, \textit{et al.} Blindly using all synthetic images for evaluation could lead to unreliable results and observations. To mitigate this issue, we propose a two-stage automatic noise filtering strategy. In the first stage, we perform sample-level filtering by assessing the overall image quality. In the second stage, we apply pixel-level filtering to eliminate noisy synthetic areas by analyzing the predicted logits from a surrogate segmentation model. 

\noindent \textbf{Sample-level Filtering.} 
In the first stage, we filter out low-quality synthetic images using CLIP-based \cite{clip} image-text similarity metric \cite{gal2022stylegan, clipscore, lance}. We denote $\mathcal{E}_I$ and $\mathcal{E}_T$ as CLIP image and text encoders, the CLIP image-text directional similarity is computed as follows:
\begin{equation}
  \rho(I, I^*, P, P^*)=  \frac{(\mathcal{E}_I(I) - \mathcal{E}_I(I^*)) \cdot (\mathcal{E}_T(P) - \mathcal{E}_T(P^*))}{|\mathcal{E}_I(I) - E_I(I^*)| \cdot |\mathcal{E}_T(P) - \mathcal{E}_T(P^*)|}
\end{equation}
$\rho$ measures consistency between image editing direction and text editing direction in high-dimensional embedding space. In addition, we also compute CLIP image-image similarity $\mathcal{E}_I(I) \cdot \mathcal{E}_I(I^*)$ to ensure image perceptual quality of synthetic images, and compute CLIP image-text similarity $\mathcal{E}_I(I^*) \cdot \mathcal{E}_T(P^*)$ to ensure achievement of target editing.

\noindent \textbf{Pixel-level Filtering.} 
However, synthetic images often contain noisy regions that either do not align with the corresponding semantic segmentation labels or include implausible content. In both cases, a pretrained segmentation model may exhibit abnormal behavior in these regions, beyond a simple decline in accuracy. Inspired by this, we propose using a segmentation model trained with real images to compare its prediction losses between the original image and the generated counterpart, and eliminate noisy regions based on a designed criterion \cite{freemask}.

Specifically, suppose that the real image set $\{I^i\}_{i=1}^{B}$ and corresponding semantic labels $\{G^i\}_{i=1}^{B}$ , we first use a semantic segmentation model to calculate its pixel-level cross-entropy loss map $\{Y^i\}_{i=1}^{B}$. Then we compute the average loss for each class $g$ as follows: 
\begin{equation}
    l_g = \sum_{i=1}^{B} \sum_{hw}^{HW} \left[ \mathbbm{1}{(G_{hw}^i = g)} \cdot Y_{hw}^i \right] \Big/ \sum_{i=1}^{B} \sum_{hw}^{HW} \mathbbm{1}(G_{hw}^i = j)
\end{equation}
where $H$ and $W$ indicates height and width, and $\mathbbm{1}(\cdot)$ is an indicator function. $l_g$ effectively represents learning difficulty for semantic class $g$ in the real image dataset, where a larger value indicates a greater challenge in segmentation.
For a generated counterpart $I$, we then compute its loss map $Y^*$. If the value of $Y^*$ at pixel location $k$ with semantic label $j$ significantly deviates from $l_g$, it is considered as noise: $Y^*_k > \alpha \cdot l_g$ or $Y^*_k < l_g / \alpha$ where $\alpha$ controls the amplitude of the loss margin. Since informative edited contents also induce large loss values, we adopt $\alpha=2$ to prevent excessive filtering.
For regions identified as noisy, we remove their corresponding pixel-level annotations from existing label maps, thereby omitting their influence on the evaluation results. Qualitative results of our localized noise are shown in the supplementary material.
\vspace{-0.5em}
\section{Dataset Analysis}
\label{sec:dataset}
\textit{What defines high-quality synthetic data for evaluation?} We argue that both realism and faithfulness to the original data are crucial. In this section, we first introduce our datasets Pascal-EA and COCO-EA generated via the \textit{Gen4Seg}. Then we compare them with previous synthetic benchmarks to analyze the quality of synthetic images.


\vspace{-0.5em}
\subsection{Pascal-EA and COCO-EA}
We utilize several popular challenging datasets including Pascal VOC \cite{pascal} and COCO-Stuff 164k \cite{cocostuff}. Pascal VOC \cite{pascal} contains 20 object categories for semantic segmentation tasks, its training and validation sets have 1,264 and 1,449 samples, respectively. COCO-Stuff 164k \cite{cocostuff} contains 172 semantic categories \textit{i.e.}, 80 thing classes, 91 stuff classes and 1 class 'unlabeled'. It includes 118k images for training, 5k images for validation. For COCO-Stuff 164k, we only use thing classes to perform local object attributes. We employ their validation sets for generation, and we final construct their counterparts \textbf{Pascal-EA}(\textbf{E}ditable \textbf{A}ttributes) and \textbf{COCO-EA}. Each has four appearance attributes subsets and two geometry attributes subsets.

\noindent \textbf{Appearance changes.} We prioritize object-centric images with simpler scenes and minimal occlusion to ensure the quality of the generated images. Thus, we discard samples without a prominent object. Specifically, for each sample in the validation sets, we first identify the most salient foreground object by ranking objects based on their area. We then select samples where the largest object occupies more than 20\% of the entire image. Finally, we extract the corresponding object mask from the segmentation map to form image-mask-label triplet for subsequent editing procedures. In text editing, we randomly choose one from multiple candidates of edited texts generated by the LLM. 

\noindent \textbf{Geometry changes.} We focus exclusively on reducing object sizes, as larger objects tend to be more salient, potentially simplifying the segmentation task. We define three levels of size reduction, with proportions [0.2, 0.4], where smaller values indicate a greater reduction in object size. For object position, we also design three levels of moving object [0.2, 0.4], where a smaller value means a smaller movement distance. The corresponding semantic maps are adjusted accordingly. In our evaluation, we focus solely on performance within the object itself, disregarding background accuracy.

The final dataset preparation involved a human-in-the-loop process for both intermediate and final cleaning, where we manually filtered the highest-quality images. Some examples are shown in Fig. \ref{fig:dataset_vis}. More details are shown in the supplementary material.

\begin{table*}[!t]
\footnotesize
\caption{Comparison of our generated images with previous benchmarks under four adverse weather conditions. ``\textit{generative}'' and ``\textit{simulator}'' refer to the images generated by generative methods and the simulator, respectively. The symbol ``-'' indicates that the benchmark does not support the required setting or functionality.}
\label{table:benchmark_compare1}
\centering
\resizebox{\textwidth}{!}{
\begin{tabular}{lc cc cc cc cc}
\toprule[1pt]
 & \multirow{2}{*}{Type} & \multicolumn{2}{c}{Snow} & \multicolumn{2}{c}{Rain} & \multicolumn{2}{c}{Fog} & \multicolumn{2}{c}{Night} \\
 \cmidrule(lr){3-4} \cmidrule(lr){5-6} \cmidrule(lr){7-8} \cmidrule(lr){9-10}
 & & CLIP Acc ($\uparrow)$ & FID ($\downarrow$)  & CLIP Acc ($\uparrow)$ & FID ($\downarrow$) & CLIP Acc ($\uparrow$) & FID ($\downarrow$) & CLIP Acc ($\uparrow$) & FID ($\downarrow$)      \\
\midrule
ACDC{\color{gray}\tiny [ICCV2021]}~\cite{acdc} & real &  1.000   &  0.000   &  1.000 & 0.000  & 1.000 &  0.000  &  1.000  &  0.000  \\
\midrule
Multi-weather{\color{gray}\tiny [ICCV2021]}~\cite{multiweather} &  \textit{generative} & 0.163 &  197.48  &  0.852   &  189.85 & - &  - &   0.906  & \textbf{154.30}   \\
Fog Cityscapes{\color{gray}\tiny [IJCV2018]}~\cite{foggydriving} & \textit{simulator} & - & - & -  & - &  0.940 & 164.00 & - &  -        \\
SHIFT{\color{gray}\tiny [CVPR2022]}~\cite{shift} &  \textit{simulator} & - &  - &  0.980   & 242.28 &  0.955  &  272.58  & 0.914 & 274.45  \\
GenVal{\color{gray}\tiny [ECCV2024]}~\cite{loiseau2024reliability} &  \textit{generative} & 0.770 &  164.59  &  0.870  & 172.04  & 0.896 & 155.09 &  0.713 & 181.57   \\
\midrule
\rowcolor{gray!40}
Ours   & \textit{generative} &   \textbf{1.000}    &  \textbf{150.61}  & \textbf{0.989}  &  \textbf{143.47}  & \textbf{0.999}  & \textbf{137.15}  & \textbf{0.965}  &  186.69  \\
\bottomrule[1pt]
\end{tabular}}
\end{table*}
\vspace{-0.5em}

\subsection{Comparison with other Datasets}

\begin{figure}
    \centering
    \includegraphics[width=\columnwidth]{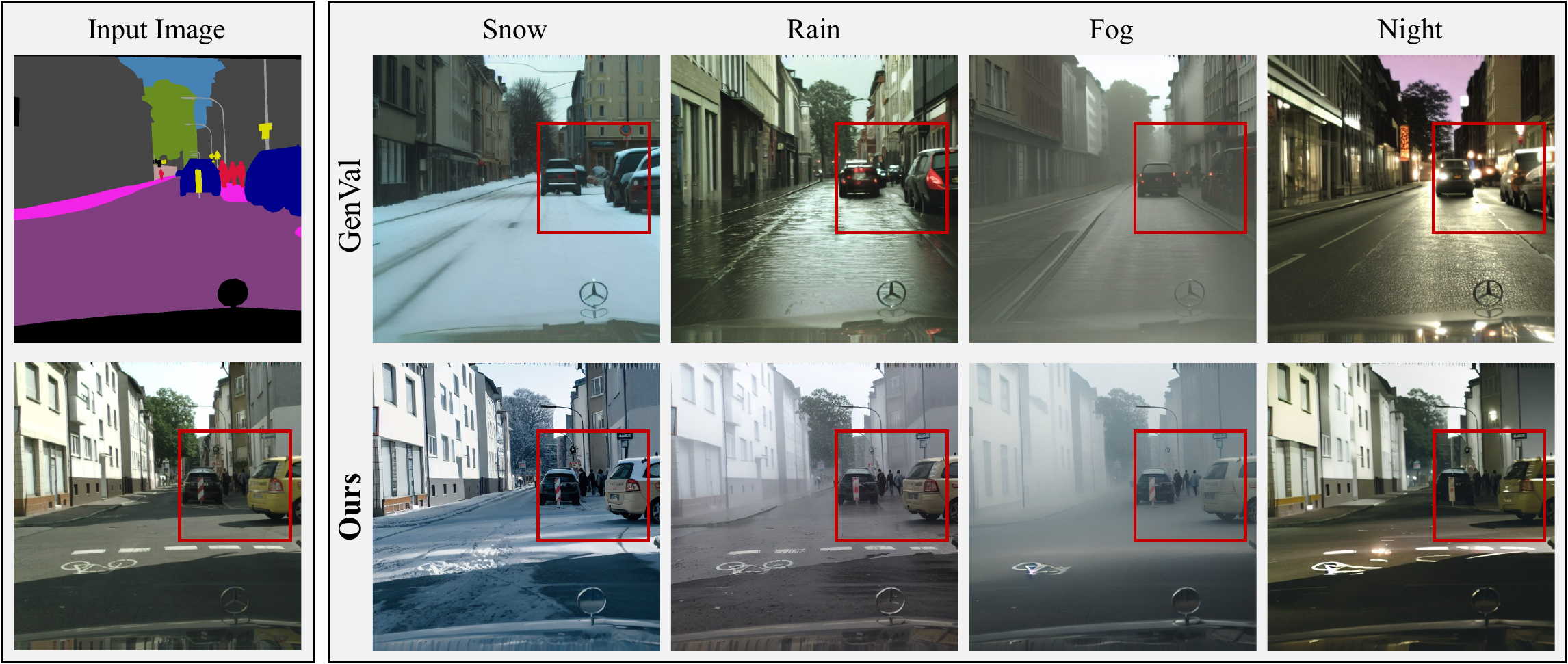}
    \caption{Qualitative Comparison of GenVal \cite{loiseau2024reliability} and Ours. We can better maintain the spatial structures of original scenes.}
    \vspace{-1.5em}
    \label{fig:benchmark_compare}
\end{figure}

\noindent \textbf{Setting.} Popular benchmarks \cite{acdc, shift, foggydriving, multiweather, loiseau2024reliability} for semantic segmentation robustness evaluation heavily focus on autonomous driving environments.
Many approaches leverage generative models \cite{multiweather, cyclegan, loiseau2024reliability} or simulators \cite{shift, foggydriving} to synthesize images under adverse weather conditions, including fog, rain, snow, and nighttime scenarios.
Following previous settings, we take the Cityscapes dataset \cite{cityscapes} as source images, and separately generate images with four adverse weathers using our proposed image appearance editing method. 
We compare our generated images with previous synthetic benchmarks including Multi-weather \cite{multiweather}, Fog Cityscapes \cite{foggydriving}, and SHIFT \cite{shift}, in terms of image realism and fidelity. The ACDC dataset \cite{acdc}, which contains normal-adverse paired samples collected from real scenes, is used as gold standard in comparison.

\noindent \textbf{Evaluation metrics.} We adopt two metrics: (1) CLIP Accuracy (CLIP Acc) \cite{clipscore}, which measures the percentage of instances where the target image achieves higher CLIP similarity to the target text than to the source text, and (2) Fréchet Inception Distance (FID), which assesses the distributional similarity between real and generated images. The FID score of synthetic data is computed with real images under normal conditions from the ACDC \cite{acdc}.

\noindent \textbf{Results.} From quantitative results illustrated in Table \ref{table:benchmark_compare1}, we can draw two observations. (1) 
In most weathers, our generated data are more realistic and distributional closer to real images, compared to previous synthetic benchmarks. But at night, our images are worse than Multi-weather \cite{multiweather} which utilizes a set of GAN and CycleGAN models \cite{vqgan, cyclegan} to transfer styles. We speculate this attributes to that diffusion models struggle in manipulating brightness and light information which is consistent with findings in \cite{yin2023cle}. (2) Comparing to GenVal \cite{loiseau2024reliability} which also uses LDM \cite{ldm} and ControlNet \cite{controlnet}, our method demonstrates superior performance.
To further investigate our differences, we visualize generated images in Fig. \ref{fig:benchmark_compare}. Notably, GenVal \cite{loiseau2024reliability} alters the original layout of the street.
This issue likely stems from that GenVal \cite{loiseau2024reliability} solely takes semantic map and text as conditions, which may not be sufficiently reliable. By incorporating additional features from the original image as an extra condition, our method can better preserve the spatial structure of the original scene. 

In essence, we demonstrate that samples generated by our pipeline have better image reality compared to conventional synthetic benchmarks. Experimental results can serve as evidence that our pipeline can be a better substitution for real images, especially as acquiring labels is laborious.

\begin{table*}[ht]
\renewcommand\arraystretch{1.05}
\centering
\caption{Benchmarking the robustness of 13 state-of-the-art methods under four appearance attributes changes on our Pascal-EA and COCO-EA datasets. The results are reported using mIoU ($\uparrow$). \textit{Recon} denotes results on reconstructed images by our \textit{Gen4Seg} without any attribute variations. We also report the average mIoU on the last row.} 
\label{tab:benchmark_app}
\resizebox{\textwidth}{!}{
\begin{tabular}{lc|ccccccc|ccccccc}
\toprule[1.28pt]
\multirow{2}{*}{Method} & \multirow{2}{*}{Backbone} & \multicolumn{7}{c|}{Pascal-EA}  & \multicolumn{7}{c}{COCO-EA}  \\ 
\cmidrule(lr){3-9} \cmidrule(lr){10-16}
    &    & Original & \textit{Recon} & Color & Material & Weather & Style & \cellcolor{gray!40} $mR$ & Original & \textit{Recon} & Color & Material & Weather & Style & \cellcolor{gray!40} $mR$ \\ 
\midrule
PSPNet{\color{gray}\tiny [CVPR2017]} \cite{pspnet}    &  ResNet50  &  77.02  &  76.10    &  72.00  &  45.14  &   69.97    &  63.83   &  \cellcolor{gray!40} 0.82 &   36.20    &  35.54     &  32.8  &  29.62     &  30.25  &  27.23    &  \cellcolor{gray!40} 0.84 \\
DeepLabV3{\color{gray}\tiny [CVPR2018]} \cite{deeplabv3}    &  ResNet50  &  75.33  &  74.26    &  71.66  &  45.49  &   69.21    &  63.7   &  \cellcolor{gray!40} 0.84 &  36.98  & 36.25 & 33.68  &      30.68 & 31.28 & 28.27 &  \cellcolor{gray!40} 0.85  \\
GCNet{\color{gray}\tiny [TPAMI2020]} \cite{gcnet}    &  ResNet50  &    76.78 &  75.53  &  70.83 &  43.09  & 69.19 & 61.79  &  \cellcolor{gray!40} 0.81  &  37.59  & 37.00  & 35.07   & 29.65 & 28.34 &    26.03 & \cellcolor{gray!40} 0.80\\
OCRNet{\color{gray}\tiny [ECCV2020]} \cite{ocr}  & HRNet48  &  78.29  &    77.00    &   73.16    &  49.37  &  73.69 & 69.08  & \cellcolor{gray!40} 0.86  & 40.55  & 39.74 & 37.69 & 33.94  & 30.04  & 29.16   &  \cellcolor{gray!40} 0.82\\
\midrule
Segmenter{\color{gray}\tiny [ICCV2021]} \cite{segmenter}  & ViT-B/16  &   82.50 & 81.19 & 78.69  &  60.29  &  78.43  & 77.04 &  \cellcolor{gray!40} 0.91  & 46.93  & 46.38 & 45.28 & 42.64  & 41.75 & 39.39  & \cellcolor{gray!40} 0.91\\
SegFormer{\color{gray}\tiny [NeurIPS2021]} \cite{segformer}    & MIT-B3 &  81.81  &  80.84  &  77.49  &  55.45  &  76.04  &  73.45 & \cellcolor{gray!40} 0.87 & 43.10  & 42.69  & 40.07 &  38.5  &  37.49  &  34.2 &  \cellcolor{gray!40} 0.87\\
MaskFormer{\color{gray}\tiny [NeurIPS2021]} \cite{maskformer}   & Swin-B &  80.49  &  79.26  &  76.91  &  54.30  &  75.61  & 71.95 &  \cellcolor{gray!40}  0.88    & 43.66  & 43.07 & 41.23 & 39.27  & 36.52  & 33.41 &  \cellcolor{gray!40} 0.87\\
Mask2Former{\color{gray}\tiny [CVPR2022]} \cite{mask2former}    & Swin-B &  88.07  &  87.52  & 85.35  & 61.52  &  85.30  & 79.09  &  \cellcolor{gray!40} 0.89  &  45.77 & 45.06  & 41.64  & 38.6  & 38.52  & 34.54 &  \cellcolor{gray!40} 0.85\\
\midrule
OVSeg{\color{gray}\tiny [CVPR2023]} \cite{ovseg} &  Swin-B  &    92.48    & 91.06 &  89.11  &  76.95   & 88.15  &  87.40  & \cellcolor{gray!40} 0.93   & 49.64  & 48.4 & 47.27 & 46.22 & 42.7 & 41.27 &  \cellcolor{gray!40} 0.92\\
ODISE{\color{gray}\tiny [CVPR2023]} \cite{odise}  &  SD-1.5  &  90.37  & 89.35  & 87.36 & 74.49 & 85.40    &  84.44  & \cellcolor{gray!40} 0.92  & 48.31 & 47.79  & 46.59 & 43.06 & 40.01 & 41.55 & \cellcolor{gray!40} 0.89\\
X-Decoder{\color{gray}\tiny [CVPR2023]} \cite{odise}    &  Focal-T  & 89.84 & 87.21 & 85.9  & 65.85     & 87.21 & 83.67 &  \cellcolor{gray!40} 0.90  & 50.12  & 49.15  & 47.46 & 44.50 & 43.85 & 41.96 & \cellcolor{gray!40} 0.90 \\
CATSeg{\color{gray}\tiny [CVPR2024]} \cite{cat}    &  Swin-B  & 94.70  & 93.68 & 91.79 & 77.86  & 91.59  & 90.39 &  \cellcolor{gray!40} 0.94  & 50.99  & 50.62 & 49.2  & 48.33 & 44.56 & 42.8 & \cellcolor{gray!40} 0.91\\
SEEM{\color{gray}\tiny [NeurIPS2023]} \cite{seem}    &  Focal-T  & 89.86 & 88.44 & 85.88 & 70.75  &   85.60 & 84.14  & \cellcolor{gray!40} 0.92  & 49.72 & 48.94 & 46.92  & 43.92  & 41.81 & 42.82 & \cellcolor{gray!40} 0.90\\
\midrule
\rowcolor{gray!40}
\multicolumn{2}{c|}{Average} & 84.43 & 83.19 & 80.47 & 60.04 & 79.64 & 76.15 & \cellcolor{gray!40} 0.89 & 44.58 & 43.89 & 41.92 & 39.15 & 37.47 & 35.59 & \cellcolor{gray!40} 0.87 \\
\bottomrule[1.28pt]
\end{tabular}}
\end{table*}

\begin{table*}[t]
\renewcommand\arraystretch{1.05}
\centering
\caption{Benchmarking the robustness of 13 state-of-the-art methods under two object geometry attribute changes on our Pascal-EA and COCO-EA datasets. The results are reported using mIoU ($\uparrow$). } 
\label{tab:benchmark_geo}
\resizebox{\textwidth}{!}{
\begin{tabular}{lc|cccccc|cccccc}
\toprule[1.28pt]
\multirow{2}{*}{Method} & \multirow{2}{*}{Backbone} & \multicolumn{6}{c|}{Pascal-EA}  & \multicolumn{6}{c}{COCO-EA}  \\ 
\cmidrule(lr){3-8} \cmidrule(lr){9-14}
&    & Original & \makecell{Size\\0.2} & \makecell{Size\\0.4} & \makecell{Position\\0.2} & \makecell{Position\\0.4} & $mR$ & Original & \makecell{Size\\0.2} & \makecell{Size\\0.4} & \makecell{Position\\0.2} & \makecell{Position\\0.4} & $mR$ \\ 
\midrule
PSPNet{\color{gray}\tiny [CVPR2017]} \cite{pspnet}    &  ResNet50  & 67.41  & 64.97  &  62.83  &  64.98  & 63.67 &  \cellcolor{gray!40} 0.95   & 22.18  &    20.89   & 19.75 &  21.44 &  20.87  & \cellcolor{gray!40} 0.93 \\
DeepLabV3+{\color{gray}\tiny [CVPR2018]} \cite{deeplabv3plus}    &  ResNet50  &  66.69  &  64.12 & 62.37 & 64.68 & 62.98 & \cellcolor{gray!40} 0.95 & 22.07  & 20.67 & 19.64  & 21.45  & 20.80  & \cellcolor{gray!40}0.94\\
GCNet{\color{gray}\tiny [TPAMI2020]} \cite{gcnet}    &  ResNet50  & 67.85 & 65.69 &  63.60 & 64.95 & 63.5  & \cellcolor{gray!40} 0.95 & 22.78 & 21.11 & 19.46 & 21.74  & 21.36 & \cellcolor{gray!40} 0.92\\
OCRNet{\color{gray}\tiny [ECCV2020]} \cite{ocr}    &  HRNet48  & 68.15  & 67.26 &    65.33   & 66.61 & 65.08 & \cellcolor{gray!40} 0.97 & 23.03 & 21.22 & 19.50 & 21.01  & 20.68 & \cellcolor{gray!40} 0.89\\
\midrule
Segmenter{\color{gray}\tiny [ICCV2021]} \cite{segmenter}& ViT-B/16 & 70.87  & 70.34   & 69.54 & 69.58  & 68.25 & \cellcolor{gray!40} 0.97 & 23.65 & 22.02 & 19.71 & 22.81  & 22.15 & \cellcolor{gray!40} 0.92\\
SegFormer{\color{gray}\tiny [NeurIPS2021]} \cite{segformer}    &  MIT-B3  &  70.25 & 69.98 & 69.06 & 68.54  & 66.74 & \cellcolor{gray!40}0.97 & 22.95 & 21.67 & 19.03 &  21.39 & 20.54 & \cellcolor{gray!40}0.90\\
MaskFormer{\color{gray}\tiny [NeurIPS2021]} \cite{maskformer}    &  Swin-B  & 65.50  & 65.19 & 63.69 & 64.55 & 63.07 & \cellcolor{gray!40}0.98  & 22.79 & 21.02  & 19.47 & 21.77  & 19.98 & \cellcolor{gray!40}0.90\\
Mask2Former{\color{gray}\tiny [CVPR2022]} \cite{mask2former}    &  Swin-B  &  77.11&  73.56 & 71.92 & 72.87 & 72.04 & \cellcolor{gray!40}0.94 & 21.81 & 19.93 & 18.73 & 20.70  & 19.64  & \cellcolor{gray!40}0.91\\
\midrule
OVSeg{\color{gray}\tiny [CVPR2023]} \cite{ovseg} &  Swin-B  & 79.46 & 78.98 &  78.43 & 79.05  & 78.35 &  \cellcolor{gray!40}0.99 & 23.42 & 21.48 & 19.82 & 22.45 & 21.54 & \cellcolor{gray!40}0.91 \\
ODISE{\color{gray}\tiny [CVPR2023]} \cite{odise} &  SD-1.5  & 78.56 & 75.71 & 71.81&  76.84 & 75.45 & \cellcolor{gray!40}0.95 & 23.45 & 21.59 & 20.07 &22.70 & 21.93 & \cellcolor{gray!40} 0.92\\
X-Decoder{\color{gray}\tiny [CVPR2023]} \cite{odise}    &  Focal-T  & 77.03  & 75.94& 73.83 & 74.89  & 74.05 & \cellcolor{gray!40}0.96 & 28.26 &25.53 & 23.81 & 26.59 & 25.58 & \cellcolor{gray!40}0.90\\
CATSeg{\color{gray}\tiny [CVPR2024]} \cite{cat}    &  Swin-B  & 80.79 & 79.44 & 78.42 & 79.15 & 77.18 & \cellcolor{gray!40}0.97 & 23.15 & 21.14 & 19.72 & 21.92 & 21.11 & \cellcolor{gray!40}0.90\\
SEEM{\color{gray}\tiny [NeurIPS2023]} \cite{seem}    &  Focal-T  & 76.24 & 75.06 &   74.65 & 75.36 & 72.60 & \cellcolor{gray!40}0.97 & 27.38  & 25.45 & 22.98 &  25.62 &  24.86 & \cellcolor{gray!40} 0.90\\
\midrule
\rowcolor{gray!40}
\multicolumn{2}{c|}{Average} & 72.76 & 71.25 & 69.65 & 70.92 & 69.45 & \cellcolor{gray!40}0.96 & 23.61 & 21.82 & 20.13 & 22.43 & 21.62 & \cellcolor{gray!40}0.91 \\
\bottomrule[1.28pt]
\end{tabular}}
\end{table*}

\section{Benchmark Results}
\label{sec:benchmark}
In this section, we present benchmark results on our Pascal-EA and COCO-EA datasets. The experiment setup, including target models and metrics, is detailed in Sec. \ref{subsec:exp_setup}. The benchmark results are shown from Sec. \ref{subsec:main_results} to \ref{subsec:data_aug} with comprehensive analysis. Finally, we explore the application of our pipeline as a data augmentation tool in Sec. \ref{subcsec:application}.

\subsection{Benchmark Setup}
\label{subsec:exp_setup}
\noindent \textbf{Target Semantic Segmentation Models.} 
We evaluate the robustness of various architectures across three categories: (1) CNN-based methods, including PSPNet \cite{pspnet}, DeepLabV3+ \cite{deeplabv3plus}, GCNet \cite{gcnet}, and OCRNet \cite{ocr}; (2) Transformer-based methods, including Segmenter \cite{segmenter}, SegFormer \cite{segformer}, MaskFormer \cite{maskformer}, and Mask2Former \cite{mask2former}; and (3) Recent open-vocabulary methods, including CATSeg \cite{cat}, OVSeg \cite{ovseg}, ODISE \cite{odise}, X-Decoder \cite{x-decoder}, and SEEM \cite{seem}. All models are implemented using MMsegmentation \cite{mmseg} and Detectron2 \cite{detectron2} toolkit. They are trained on the original training sets of Pascal VOC \cite{pascal} and COCO Stuff \cite{cocostuff}. For comparison, we use the officially released weights and follow the original training recipes to ensure optimal performance. 

\noindent \textbf{Target Data Augmentation Methods.} 
We also investigate the robustness of the segmentation model with different data augmentation algorithms. We use vanilla Mask2Former \cite{mask2former} as a baseline model, and select several representative methods: CutOut \cite{cutout}, CutMix \cite{cutmix}, AugMix \cite{augmix}, and a method from Hendrycks \textit{et. al.} \cite{weather}. CutOut \cite{cutout} is initially proposed in image classification for a more robust representation by randomly masking out square patches from an image. CutMix \cite{cutmix} encourages to learn from multiple objects in a single image by replacing a patch with content from another sample. AugMix \cite{augmix} combines diverse augmentations (e.g., rotations, translations, and color jitter) in a stochastic and compositional manner. Hendrycks \textit{et. al.} \cite{weather} similarly combines blur, fog, and impulse noise effects.

\noindent \textbf{Evaluation metric.} 
To intuitively demonstrate the robustness of target methods, in each type of attribute variations, we use the mean performances on the edited subset and the relative degradation compared to the reconstructed original data as our main evaluation metrics.
Specifically, we adopt mIoU(\%), \textit{i.e.}, mean Intersection over Union on each class as our metric. First, we compute mIoU on reconstructed original data to eliminate the effects of the diffusion process on segmentation performances. Next, we calculate the average mIoU across all edited subsets, which we denote as R(Robustness)mIoU, and the averaged relative performance degradation is computed as follows:
\begin{equation}
    mR = \text{RmIoU / mIoU} 
\end{equation}
In this way, $mR$ indicates the overall robustness to certain attribute variations.

\begin{table*}[t]
\centering
\caption{mIoU ($\uparrow$) of four augmentation algorithms on our Pascal-EA. The best results are in bold.}
\label{tab:benchmark_aug}
\resizebox{\linewidth}{!}{
\begin{tabular}{l cccccc cccccc}
\toprule[1pt]
\multirow{2}{*}{Method} & \multicolumn{6}{c}{Appearance}  & \multicolumn{6}{c}{Geometry} \\
  \cmidrule(lr){2-7}  \cmidrule(lr){8-13}  
& \textit{Recon} & Color & Material & Weather & Style & $mR$ & Original & \makecell{Size\\0.2} & \makecell{Size\\0.4} & \makecell{Position\\0.2} & \makecell{Position\\0.4} & $mR$ \\
\midrule
Mask2Former{\color{gray}\tiny [CVPR2022]} \cite{mask2former} & 87.52 & 85.35 & \textbf{61.52} & 85.30 & 79.09 & \cellcolor{gray!40} \textbf{0.89} & 77.11 & 73.56 & 71.92 & 72.87 & 72.04 & \cellcolor{gray!40}0.94    \\
\midrule
+CutOut{\color{gray}\tiny [Arxiv2017]} \cite{cutout}   & 89.56 & 84.20 & 48.47 & 85.17 & 75.68 & \cellcolor{gray!40} 0.82 & 76.82 & 74.98 & 72.44 & 72.58 & 71.61  & \cellcolor{gray!40} 0.94\\
+CutMix{\color{gray}\tiny [ICCV2019]} \cite{cutmix}   & 90.64 & 86.44 & 52.42 & 84.34 & 77.51 & \cellcolor{gray!40} 0.83 & \textbf{77.56} & \textbf{76.62} & \textbf{74.66} & \textbf{77.01} & \textbf{74.47}  & \cellcolor{gray!40} \textbf{0.98}\\
+Hendrycks \textit{et. al.}{\color{gray}\tiny [ICLR2019]} \cite{weather}  & 91.19 & 83.27 & 55.63 & 86.21 & 80.05 & \cellcolor{gray!40} 0.84 & 75.90 & 72.45 & 70.89 & 71.14 & 68.93 &   \cellcolor{gray!40} 0.93\\
+AugMix{\color{gray}\tiny [ICLR2020]} \cite{augmix}   & \textbf{91.58} & 85.54 & 56.79 & \textbf{87.03} & \textbf{80.77} & \cellcolor{gray!40} 0.84 & 75.71 & 73.9 & 71.25 & 71.33 & 69.38  & \cellcolor{gray!40} 0.94\\
\bottomrule[1pt]
\end{tabular}}
\vspace{-0.2em}
\end{table*}

\vspace{-0.2em}
\subsection{Semantic Segmentation Models Evaluation} 
\label{subsec:main_results}
Benchmark results are reported in Tab. \ref{tab:benchmark_app} and Tab. \ref{tab:benchmark_geo}, in which CNN-based, Transformer-based, and open vocabulary methods are demonstrated in upper, median, and lower parts, respectively. Some qualitative results are shown in the supplementary material. Based on the results, we summarize the following findings:

\textbf{Finding-1:} \textit{Diffusion process induces subtle disturbances on segmentation performances which are negligible compared to the impact of attribute variations.}

To verify whether the diffusion process affects the reliability of our evaluations, we create a reference set by reconstructing the original validation set, referred to as Recon in Tab. \ref{tab:benchmark_app}. This process involves adding noise to the original images and then denoising them using non-edited texts. In the last row of Tab. \ref{tab:benchmark_app}, segmentation methods achieve 84.83\% and 44.58\% mIoU on Pascal-EA and COCO-EA, and 83.19\% and 43.89\% on the reconstructed versions. The average performance drop is only 1.24\% and 0.69\%, respectively. In comparison, performances on other appearance attribute variations fall below 81\% and 41\%, demonstrating that the impact of attribute variations is much greater than that of the diffusion process.

\textbf{Finding-2:} \textit{All models are more vulnerable to material changes than to other appearance attributes adjustments.}

The last row in Tab. \ref{tab:benchmark_app} shows that the average mIoU performance under local material changes is 60.04\% and 39.15\%, while for local color variations, the values are 80.47\% and 41.92\% on Pascal-EA and COCO-EA, respectively. For global style and weather changes, the values are 79.64\%/76.15\% and 37.47\%/35.59\% on Pascal-EA and COCO-EA, respectively. These results indicate that object material variation has a significantly greater impact on segmentation performance than other attributes. 
We speculate that this is because material alteration affects the visual texture of objects, and segmentation methods are particularly sensitive to textures. Recent work \cite{trapped} also shows similar results. And our observation aligns with findings from classification models, which exhibit texture bias, as highlighted in \cite{texture-bias}.

\textbf{Finding-3:} \textit{Under appearance variations, Transformer-based methods are more robust than CNN-based methods, and recent open-vocabulary methods show the most superior robustness.}

Firstly, in Tab. \ref{tab:benchmark_app}, Transformer-based methods like Segmenter \cite{segmenter}, SegFormer \cite{segformer}, MaskFormer \cite{maskformer}, and Mask2Former \cite{mask2former} achieve 0.91, 0.87, 0.88, and 0.89 metrics of \textit{mR} on Pascal-EA, respectively. These results are higher than those of CNN-based methods on both Pascal-EA and COCO-EA, supporting the observation that recent Transformer-based methods are more reliable than traditional CNN-based models. This finding is consistent with previous research \cite{de2023reliability}. The reasons are multifaceted, and we speculate from three perspectives: 1) Architectural inductive bias: the self-attention mechanism in vision transformers \cite{attention} enables direct modeling of long-range and global context \cite{vit}, unlike the strictly local receptive fields of standard CNNs. 2) Pre-training scale: vision transformers are often pre-trained on a larger pre-training scale (datasets, duration), which enhances knowledge transfer and out-of-distribution generalization, as observed in \cite{paul2022vision}. (3) Model scale: transformer backbones typically have more parameters and greater compute budgets, which systematically improve representation quality and robustness \cite{zhai2022scaling,dehghani2023scaling}.

Secondly, in open-vocabulary frameworks, all algorithms achieve scores above 0.90 in \textit{mR} on both Pascal-EA and COCO-EA, significantly outperforming close-set methods. This is attributed to their stronger backbones, hybrid-representation architectures incorporating CLIP \cite{clip}, and massive training data. Notably, the ODISE \cite{odise} model shows inferior performance compared to CATSeg \cite{cat} and OVSeg \cite{ovseg}. Since the main difference lies in the mask proposal backbone, we argue that Stable Diffusion (SD-1.5) \cite{ldm} exhibits relatively worse robustness compared to other specialized backbones, such as Focal-T and Swin-B, in segmentation tasks.

\textbf{Finding-4:} \textit{All types of models exhibit comparable robustness to geometry attribute variations.}

The Tab. \ref{tab:benchmark_geo} illustrates that the \textit{mR} discrepancy between Transformer-based and CNN-based methods diminishes when considering geometry variations. For instance, Mask2Former \cite{mask2former} and OCRNet \cite{ocr} achieve comparable scores of 0.94 and 0.95 on Pascal-EA. Similarly, recent open-vocabulary methods attain around 0.95 on Pascal-EA and 0.90 on COCO-EA. In contrast to the findings in appearance variations, all types of methods demonstrate comparable robustness under geometry changes, with some advanced models, like SEEM \cite{seem}, even performing worse than baseline models on COCO-EA. For example, SEEM \cite{seem} records 27.38\% and 24.86\% mIoU, while DeepLabv3 \cite{deeplabv3} achieves much lower values 22.07\% and 20.80\% under the original and Position-0.4 sets, on COCO-EA, respectively, yet SEEM's \textit{mR} value is 0.90, which is lower than DeepLabv3's 0.94.
These results reveal that better in-distribution performance may predict higher accuracy under geometry shifts. This observation aligns with the findings in ImageNet-E \cite{li2023imagenet}, which studies network design effectiveness under attribute shifts. Our results complement this by highlighting that more powerful architectures and extensive training data do not guarantee greater robustness.

\begin{table*}[t]
\centering
\caption{Comparison of different methods from Cityscapes \cite{cityscapes} (source) to ACDC \cite{acdc} (target) using the mIoU ($\uparrow$) metric. The results are reported on the Cityscapes validation set, four individual scenarios of ACDC, and the average (Avg). $^\dagger$ means results are directly from \cite{issa}. \textit{Oracle} indicates the supervised training, serving as performance upper bounds.}
\label{tab:application2}
\resizebox{\linewidth}{!}{
\begin{tabular}{l|c|ccccc|c|ccccc}
\toprule
& \multicolumn{6}{c|}{HRNet \cite{hrnet}} & \multicolumn{6}{c}{Segformer \cite{segformer}} \\ 
Method & Cityscapes & Rain & Fog & Snow & Night & Avg. & Cityscapes & Rain & Fog & Snow & Night & Avg.\\ 
\midrule
Baseline$^\dagger$  & 70.47 & 44.15 & 58.68 & 44.20 & 18.90 & 41.48  & 67.90 & 50.22 & 60.52 & 48.86 & 28.56 & 47.04\\ 
\midrule
CutOut{\color{gray}\tiny [Arxiv2017]} \cite{cutout}   & 71.39 & 40.29 & 57.70 & 43.98 & 16.55 & 39.63  & 68.93 & 47.68 & 60.34 & 46.98 & 26.49 & 45.37     \\
CutMix$^\dagger${\color{gray}\tiny [ICCV2019]} \cite{cutmix}   & 72.68 & {42.48} & {58.63} & 44.50 & {17.07} & {40.67} & {69.23} & {49.53} & 61.58 & {47.42} & {27.77} & {46.57} \\
Hendrycks \textit{et. al.}$^\dagger${\color{gray}\tiny [ICLR2019]} \cite{weather}  & 69.25 & \textbf{50.78} & 60.82 & {38.34} & 22.82 & 43.19  & 67.41 & 54.02 & 64.74 & 49.57 & 28.50 & 49.21 \\
StyleMix$^\dagger${\color{gray}\tiny [CVPR2021]}\cite{stylemix} & 57.40 & {40.59} & {49.11}  & {39.14} & 19.34 & {37.04} & 65.30 & 53.54 & 63.86 & 49.98 & 28.93 & 49.08\\ 
ISSA$^\dagger${\color{gray}\tiny [WACV2023]}\cite{issa} & 70.30 & {50.62} & \textbf{66.09}  & \textbf{53.30} & \textbf{30.18} & \textbf{50.05} & 67.52 & \textbf{55.91} & \textbf{67.46} & \textbf{53.19} & \textbf{33.23} & \textbf{52.45} \\ 
\rowcolor{gray!40}
\textbf{Ours}    & 65.77 & 46.40 & \underline{61.62} & \underline{49.88} & \underline{28.59} & \underline{46.67} & 63.48 & \underline{52.20} & \underline{65.78} & \underline{51.40} & \underline{30.22} & \underline{51.04}   \\
\midrule
\textit{Oracle}$^\dagger$ & - & 65.67 & 75.22 & 72.34 & 50.39 & 65.90  & - & 63.67 & 74.10 & 67.97 & 48.79 & 63.56 \\
\bottomrule
\end{tabular}}
\end{table*}

\begin{table}
\renewcommand\arraystretch{1.05}
\centering
\caption{The results of different data augmentation techniques on Pascal VOC \cite{pascal} and COCO-Stuff 164k \cite{coco}.}
\label{tab:application1}
\resizebox{\linewidth}{!}{
\begin{tabular}{ccccccc}
\toprule[1pt]
 & Mask2Former & +CutOut & +CutMix  & +AugMix & \cellcolor{gray!40} +\textbf{Our}\\
\midrule
Pascal VOC & 87.52 & 89.56 & 90.64 & 91.58 & 91.77 \cellcolor{gray!40} \\
COCO Stuff & 45.06 & 45.98 & 46.52 & 46.84 & 47.03 \cellcolor{gray!40} \\
\bottomrule[1pt]
\end{tabular}}
\end{table}

\subsection{Data Augmentation Methods Evaluation}
\label{subsec:data_aug}
It is also important to systematically investigate how different classical data augmentation strategies perform under variations in appearance and geometry attributes. The results of four augmentation methods are presented in Tab. \ref{tab:benchmark_aug}. We can obtain several findings: 

\textbf{Finding-5:} \textit{CutMix improves model robustness against geometry variations while other methods fail.}

As illustrated in the table, after applying CutMix augmentation, there are significant improvements in all geometry variations. For instance, mIoU increases from 73.56\% to 76.62\% and from 72.87\% to 77.01\% on the Size-0.2 and Position-0.2 subsets, respectively, with a corresponding 0.04 improvement in the \textit{mR} value. However, other data augmentation methods show limited robustness, with segmentation accuracy experiencing considerable declines, particularly with larger variation strengths. For example, the mIoU of AugMix \cite{augmix} drops from 72.04\% to 68.93\% and from 71.92\% to 71.25\% on the Position-0.4 and Size-0.4 subsets.
In addition, CutMix \cite{cutmix} improves mIoU values from 77.11\% to 77.56\% on the original validation set while other methods reduce it to 76.82\%, 75.90\%, and 75.51\%, respectively. 
The reason behind is: other augmentations alter appearance. For example, Cutout \cite{cutout} randomly masks input images, Hendrycks \textit{et. al.} \cite{weather} adds color jitter or noise. In contrast, CutMix \cite{cutmix} replaces a rectangular region with pixels from another image and mixes the labels, thereby exposing the segmenter to partial and displaced object evidence while maintaining consistent supervision for those partial views. The original CutMix study \cite{cutmix} reports improved weakly-supervised localization, which aligns with our finding.

\textbf{Finding-6:} \textit{All data augmentation methods fail to improve robustness against appearance variations.}

While all augmentation methods improve in-distribution performances, they all fail to improve robustness against appearance variations. For instance, AugMix \cite{augmix} and Hendrycks \textit{et al.} \cite{weather} increase mIoU on the reconstructed set from 87.52\% to 91.58\% and 91.19\%, respectively, but their \textit{mR} score declines from 0.89 to 0.84.
Furthermore, Hendrycks \textit{et. al.} \cite{weather} and AugMix \cite{augmix} exhibit opposite trends under object-level and image-level appearance variations. They improve mIoU value on global-level transfer,\textit{i.e.} weather and style, to 86.21\%/87.03\% to 80.05\%/80.77\%, respectively. However, they struggle with object-level variations, with a drop from 61.52\% to 55.63\%/56.79\% on the material set.
We speculate that local object variations alter visual content, while global variations shift visual styles. Since Hendrycks \textit{et al.} \cite{weather} and AugMix \cite{augmix} manipulate style-related global image features like brightness, models trained on these methods show improvement only in global variations. This highlights that these methods are insufficient to address both object-level and image-level changes. This may serve as evidence that our benchmark is meaningful for comprehensive evaluations.

It is important to highlight that our \textit{Gen4Seg} framework is not only a benchmark but can also serve as a data augmentation tool to enhance the robustness of semantic segmentation models. Detailed results and discussions on its effectiveness as a data augmentation strategy are provided in the supplementary material.

\subsection{Applications} 
\label{subcsec:application}
Our proposed \textit{Gen4Seg} is primarily developed as a diagnostic benchmark. At the same time, we also explore its potential as a data augmentation tool to examine whether the generated samples can provide auxiliary benefits for improving segmentation robustness. We emphasize that data augmentation is not our main objective, since the pipeline is deliberately designed to be optimization-free and does not involve any task-specific augmentation strategies or tuning.

To this end, we further evaluate the use of \textit{Gen4Seg} for augmenting training data under two representative settings: (1) in-distribution testing on Pascal VOC \cite{pascal} and COCO-Stuff 164k \cite{coco} original validation set; (2) domain generalization in which models are trained on Cityscapes \cite{cityscapes} and tested on samples with adverse weather conditions (fog, snow, rain, and night) on the ACDC dataset \cite{acdc}. The results are presented in Tab. \ref{tab:application2} and Tab. \ref{tab:application1}. 

In Tab. \ref{tab:application1}, we mix our generated appearance variations and geometry variations with a 50/50 ratio, and mix generated samples with real samples with a 50/50 ratio. The results illustrate that in-distribution performance of Mask2Former \cite{mask2former} improves from 84.52\% to 91.77\% and from 45.06\% to 47.03\% on Pascal VOC and COCO-Stuff 164k, respectively.
In Tab. \ref{tab:application2}, we use our \textit{Gen4Seg} to generate four weather scenarios and follow the training protocol of ISSA \cite{issa}. The results indicate that while CutOut \cite{cutout} and CutMix \cite{cutmix} improve in-distribution performance, they suffer performance drops under all out-of-distribution weather conditions. This can be attributed to the fact that these methods focus on changing local visual cues, making them more vulnerable to global style shifts. 
Moreover, Hendrycks \textit{et al.} \cite{weather} show a performance decline in snow scenarios, and StyleMix \cite{stylemix} experiences degradation across all weather conditions. We attribute this to poor synthetic image quality. 

ISSA \cite{issa} performs the best, showing consistent improvements in mIoU across all four weather scenarios. Our method remains competitive and outperforms other methods, except for ISSA \cite{issa}. ISSA \cite{issa} leverages StyleGAN-based inversion \cite{stylegan}, whose well-structured latent space facilitates clean content–style disentanglement and exemplar-driven style mixing. Our diffusion-based, text-guided style transfer is more flexible and yields diverse variants, but it does not offer the same explicit low-dimensional content–style factorization. Note that our method is designed to generate challenging test data for evaluating segmentation models, not for augmentation tasks. As a result, the augmentation results of our method are not optimized for augmentation, yet our method still outperforms several augmentation baselines, demonstrating the auxiliary abilities of our method in other applications.



\section{Gen4Seg Pipeline Analysis}
In this section, we analyze our proposed appearance and geometry editing methods by comparing them with existing diffusion-based approaches, and then validating our designs by ablation studies.

\vspace{-0.2em}
\subsection{Comparison with other Editing Methods}
\label{subsec:compare_diffusion}

\begin{figure}[t]
    \centering
    \includegraphics[width=\columnwidth]{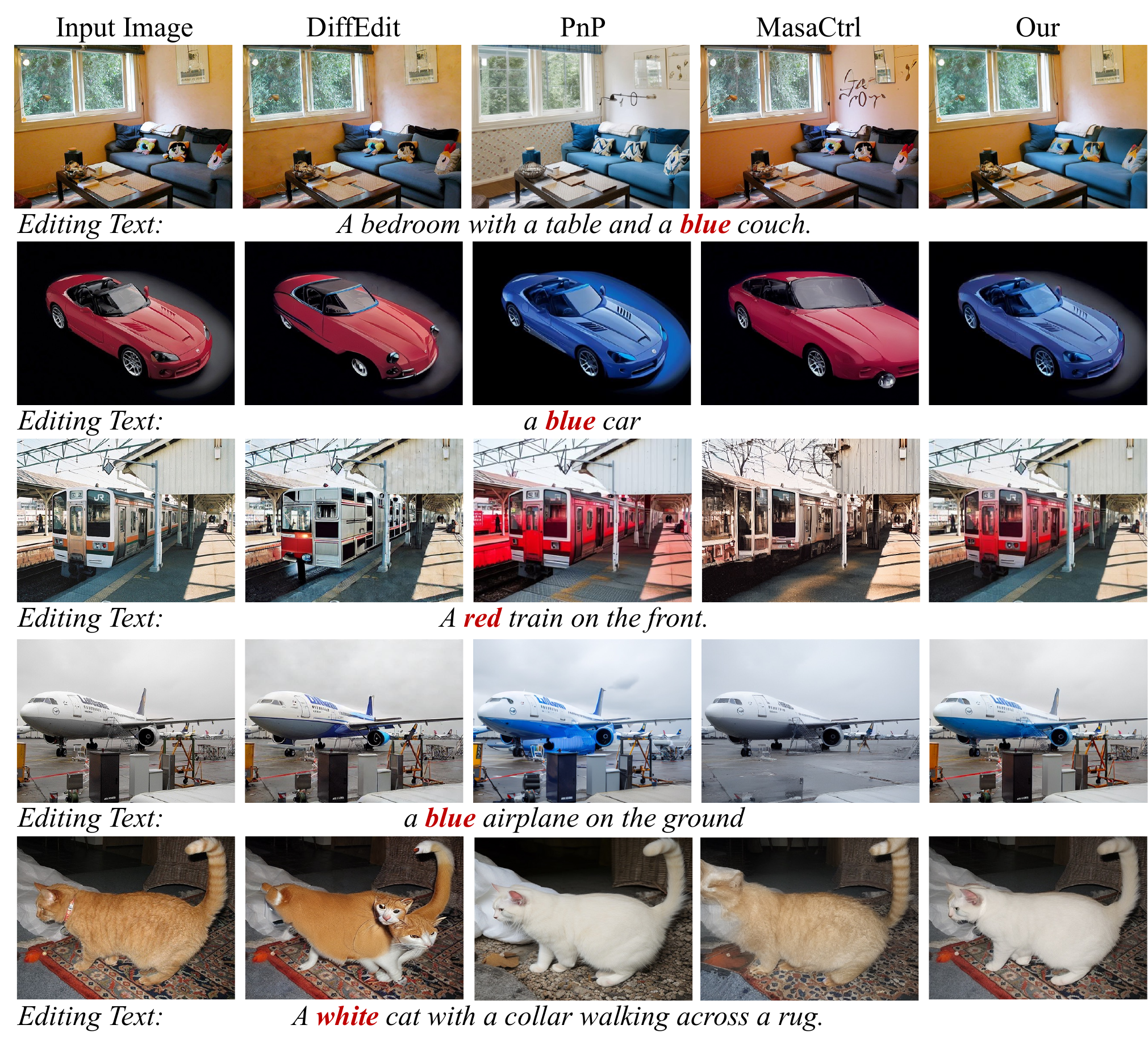}
    \caption{Qualitative results of our method and existing text-driven image editing methods.} 
    \vspace{-1em}
    \label{fig:edits_compare}
\end{figure}

\noindent \textbf{Appearance Editing Methods.} 
To demonstrate the superiority of our method, we compare our Mask-Guidance Diffusion method (introduced in Sec. \ref{subsubsec:mask_diffusion}) with state-of-the-art diffusion-based image editing algorithms, focusing on the preservation of the object’s internal structure and the avoidance of disruption to irrelevant background areas. 

We adopt Masked-LPIPS and DINO Distance (DINO-Dist) \cite{dinodist}, which assesses perceptual structural differences in the intended editing area. Comparison methods include InstructPix2Pix \cite{instructpix2pix}, PnP \cite{pnp}, Prompt-to-Prompt (P2P) \cite{p2p}, and MasaCtrl \cite{masactrl}, DiffEdit \cite{diffedit} and Blended Latent Diffusion (BLD) \cite{blended}. 
Evaluations are conducted on tasks involving object color and material replacement. The results are reported in Tab. \ref{tab:edits_compare} and Fig. \ref{fig:edits_compare}. 

We can obtain the following observations. 
Firstly, previous approaches exhibit high scores in both DINO-Dist and Masked-LPIPS, but the underlying reasons differ. Since DiffEdit \cite{diffedit} and BLD \cite{blended} intend to repaint a new object according to editing text using a mask to specify the edited region, they can ensure that irrelevant background information is not affected.
But they do not achieve fine-grained control on object inner structures, \textit{e.g.} editing color will change other core properties (as shown in Fig. \ref{fig:edits_compare}).  PnP \cite{pnp} and MasaCtrl \cite{pnp} act differently: they achieve precise control of the object's inner structural consistency but will disturb details of the adjacent background (as shown in Fig. \ref{fig:edits_compare}). Therefore, these approaches could impede the reliability of the generated images for evaluation.
Secondly, our method achieves the lowest scores 0.004 in DINO-Dist and 0.154 in Masked-LPIPS as shown in Tab. \ref{tab:edits_compare}, which implies that our method can better maintain object inner structures.
By introducing extra spatial constraints to optimize noise latent in the diffusion forward process, we achieve the best performances in changing object appearance attributes without affecting other information in both simple and complex scenarios, as shown in Fig. \ref{fig:edits_compare}.

In summary, we prove that existing editing approaches are not adequate as an evaluator for segmentation models, and demonstrate the superiority of our proposed method in terms of the faithfulness of synthetic images. 

\begin{table}[t]
\centering
\caption{Quantitative results of existing image editing models. We compare them with our method by editing object color and material on the Pascal VOC \cite{pascal} dataset. MGD is the conference version of our method.} 
\label{tab:edits_compare}
\resizebox{\columnwidth}{!}{
\begin{tabular}{lcccc}
\toprule[1.28pt]
\multirow{3}{*}{Method}  & \multicolumn{2}{c}{\textbf{Material}} & \multicolumn{2}{c}{\textbf{Color}} \\ 
\cmidrule(lr){2-3}  \cmidrule(lr){4-5} 
& \makecell{DINO \\ Dist ($\downarrow$) } &  \makecell{Masked \\ LPIPS ($\downarrow$)}  & \makecell{DINO \\ Dist ($\downarrow$) } &  \makecell{Masked \\ LPIPS ($\downarrow$)} \\
\midrule
BLD{\color{gray}\tiny [SIGGRAPH2023]}~\cite{blended}           &  0.081  & 0.413          &  0.092  & 0.384  \\
DiffEdit{\color{gray}\tiny [ICLR2023]} \cite{diffedit}         &  0.068  & 0.407          &  0.061  & 0.372  \\
\midrule
InstructPix2Pix{\color{gray}\tiny [CVPR2022]}~\cite{instructpix2pix} &  0.059  & 0.393          &  0.043  & 0.359  \\
P2P{\color{gray}\tiny [CVPR2023]}~\cite{p2p}            & 0.038   & 0.255     &  0.014  & 0.272  \\
PnP{\color{gray}\tiny [CVPR2023]}~\cite{pnp}            & 0.047   & 0.329     &  0.021  & 0.295  \\
MasaCtrl{\color{gray}\tiny [ICCV2023]}~\cite{masactrl}  & 0.054   & 0.360     &  0.036  & 0.308  \\
MGD {\color{gray}\tiny [CVPR2024]}~\cite{yin2024benchmarking} & 0.007 & 0.189 & 0.005 & 0.196 \\
\midrule
\rowcolor{gray!40}
\textbf{Ours} & \textbf{0.004}  & \textbf{0.154}  &  \textbf{0.002}  & \textbf{0.108}  \\
\bottomrule[1.28pt]
\end{tabular}}
\end{table}

\begin{figure}[ht]
    \centering
    \includegraphics[width=\columnwidth]{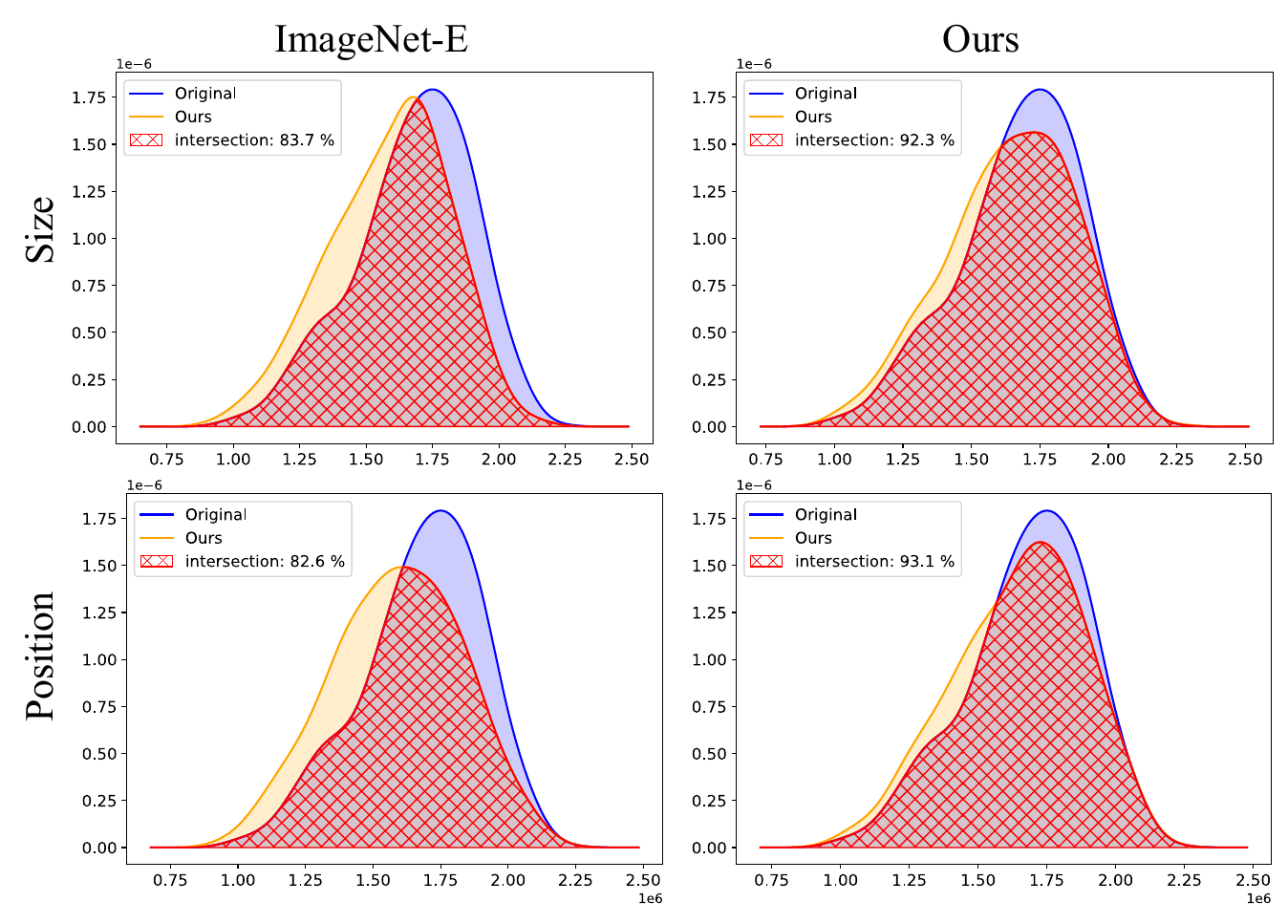}
    \caption{Distributions measured by GradNorm \cite{gradnorm} of in-distribution images (Pascal VOC \cite{pascal} dataset) and counterparts with object size and position changes, generated by our method and the ImageNet-E \cite{li2023imagenet}.} 
    \vspace{-1em}
    \label{fig:edits_compare_geo}
\end{figure}

\noindent \textbf{Geometry Editing Methods.} 
It is crucial that the geometry editing method does not introduce out-of-distribution content, which could compromise the reliability of the evaluation results. Therefore, our goal is to generate in-distribution data with minimal distributional gap from the original images. We compare our proposed geometry editing approach introduced in Sec. \ref{subsec:geo_edit}) with the previous work, ImageNet-E \cite{li2023imagenet}.
Following the previous setting \cite{li2023imagenet}, we employ the distributional shift detection method, GradNorm \cite{gradnorm}, to quantify the deviation of our generated images from their original distribution. 
Additionally, we use the Pascal VOC validation set \cite{pascal} as the source images, where selected objects are randomly resized and repositioned.

The results are presented in Fig. \ref{fig:edits_compare_geo}. The x-axis represents the distribution of scores predicted by GradNorm \cite{gradnorm}, while the y-axis denotes the frequency of these scores.  A larger intersection area indicates a closer approximation to the original distribution, leading to improved background inpainting quality and overall editing fidelity. Our method achieves an intersection (red) of 92.3\% and 93.1\% for size and position changes, respectively. Editing tool from ImageNet-E \cite{li2023imagenet} gets 83.7\% and 82.6\% intersection.

We attribute this improvement to the different inpainting techniques used in our method compared to theirs. In contrast to ImageNet-E \cite{li2023imagenet}, which involves two diffusion steps: object removal through inpainting followed by copy-paste fusion, our method has a single diffusion step: direct copy-paste and soft-mask diffusion repainting. We speculate that completely erasing an object via diffusion may introduce more instability in the generated content. Since our method inpaints a smaller area (most likely in object boundaries, seen in Fig. \ref{fig:geo_edit}) compared to ImageNet-E \cite{li2023imagenet}, it results in better generation performance.

\begin{figure}[t]
    \centering
    \includegraphics[width=\columnwidth]{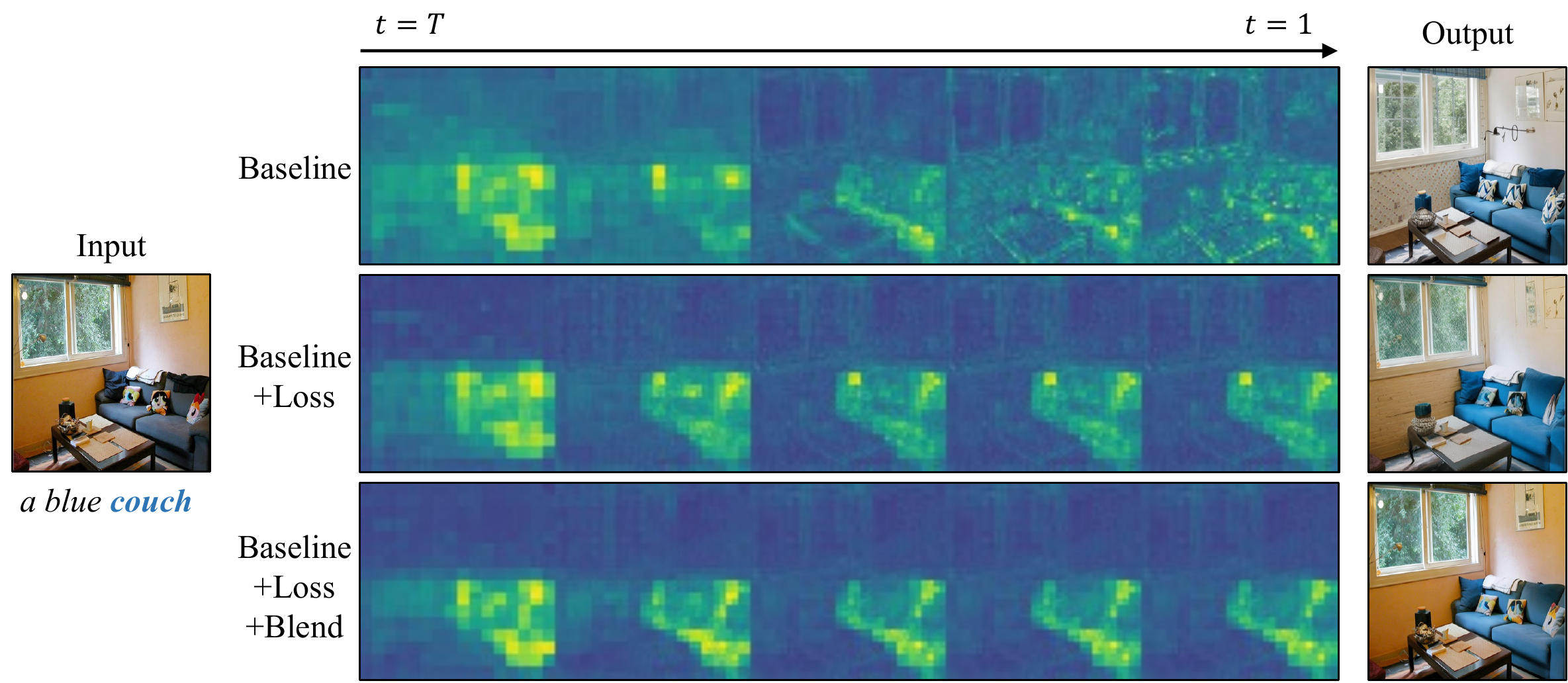}
    \caption{Visualization of cross-attention maps. There is considerable improvement after applying our strategies.}
    \vspace{-1.5em}
    \label{fig:ablation_loss}
\end{figure}

\subsection{Ablation Study}

\noindent \textbf{Appearance Attributes Editing.} 
We first demonstrate the indispensable role of our proposed Mask-Guidance Energy Function (denoted as \textit{Loss} here for brevity, Eq. \ref{eq:energy_function}) and blend operation (Eq. \ref{eq:blend}). In Tab. \ref{tab:ablation_app}, we perform localized editing of object color, material, and textures. The baseline without our strategies shows a considerable performance gap. By incorporating the Mask-Guidance Energy Function, the results improve substantially (DINO Dist decreases from 0.052 to 0.036, and CLIP Accuracy increases from 0.59 to 0.70). The blend operation further enhances the results. When both strategies are combined, the performance reaches 0.005 in DINO Distance and 0.74 in CLIP Accuracy, significantly outperforming the baseline.
To further investigate the behaviors of our strategies, we visualize the mean cross-attention maps of the denoising U-Net at different timesteps in Fig. \ref{fig:ablation_loss}. We observe that the baseline struggles with object localization (in this case, a “couch”) during the later steps, which leads to substantial alterations in background information. Introducing the Mask-Guidance Energy Function greatly improves localization accuracy. However, it still fails to entirely suppress activation in the background, causing minor changes to the background content. The combination of the blend operation with the Mask-Guidance Energy Function yields the best performance.

\begin{table}[h]
\renewcommand\arraystretch{0.9}
\centering
\caption{Ablation results of appearance attributes editing.}
\label{tab:ablation_app}
\resizebox{\columnwidth}{!}{
\begin{tabular}{ccccc}
\toprule[0.8pt]
\multicolumn{3}{c}{Our Strategies} & \multirow{2}{*}{\makecell{DINO \\ Dist ($\downarrow$) }} &  \multirow{2}{*}{\makecell{CLIP \\ Acc ($\uparrow$)}}  \\
\cmidrule(lr){1-3}
\textit{Loss} (Eq. \ref{eq:energy_function}) & \textit{Blend} (Eq. \ref{eq:blend}) & ControlNet \\
\midrule
 &  &  & 0.052 & 0.59 \\
\checkmark & & & 0.036 & 0.62 \\
& \checkmark &  & 0.010 & 0.63 \\
&  & \checkmark & 0.048 & 0.61 \\
\checkmark & \checkmark & & 0.007 & 0.70 \\
\rowcolor{gray!40}
\checkmark & \checkmark & \checkmark & \textbf{0.005} & \textbf{0.74} \\
\bottomrule[0.8pt]
\end{tabular}}
\end{table}

\begin{table}[h]
\renewcommand\arraystretch{0.9}
\centering
\caption{Ablation results of geometry attributes editing.}
\label{tab:ablation_geo}
\resizebox{\columnwidth}{!}{
\begin{tabular}{ccccc}
\toprule[0.8pt]
\multicolumn{2}{c}{Our Strategies} & \multirow{2}{*}{FID ($\downarrow$)} &  \multirow{2}{*}{LPIPS ($\downarrow$)}  & \multirow{2}{*}{\makecell{DINO \\ Dist ($\downarrow$) }} \\
\cmidrule(lr){1-2}
VLM \cite{llava} & ControlNet \cite{controlnet} & & \\
\midrule
 &  & 36.08 & 0.478 & 0.045\\
\checkmark & & 27.92 & 0.334 & 0.022\\
& \checkmark & 27.55 & 0.346 & 0.028\\
\rowcolor{gray!40}
\checkmark & \checkmark & \textbf{24.60}  & \textbf{0.295} & \textbf{0.016} \\
\bottomrule[0.8pt]
\end{tabular}}
\end{table}

\noindent \textbf{Geometry Attributes Editing.} 
In Tab. \ref{tab:ablation_geo}, we randomly move the object while maintaining its whole body inside the image. The baseline without our strategies repaints the image with prompt ``nothing'', and shows considerable performance gaps. By incorporating the VLM guidance and ControlNet step-by-step, the results on FID, LPIPS and DINO Dist decrease from 36.08 to 24.60, from 0.478 to 0.295 and from 0.045 to 0.016, respectively. Thus, we validate the effectiveness of the VLM and ControlNet guidance.

\begin{figure}[t]
    \centering
    \includegraphics[width=\linewidth]{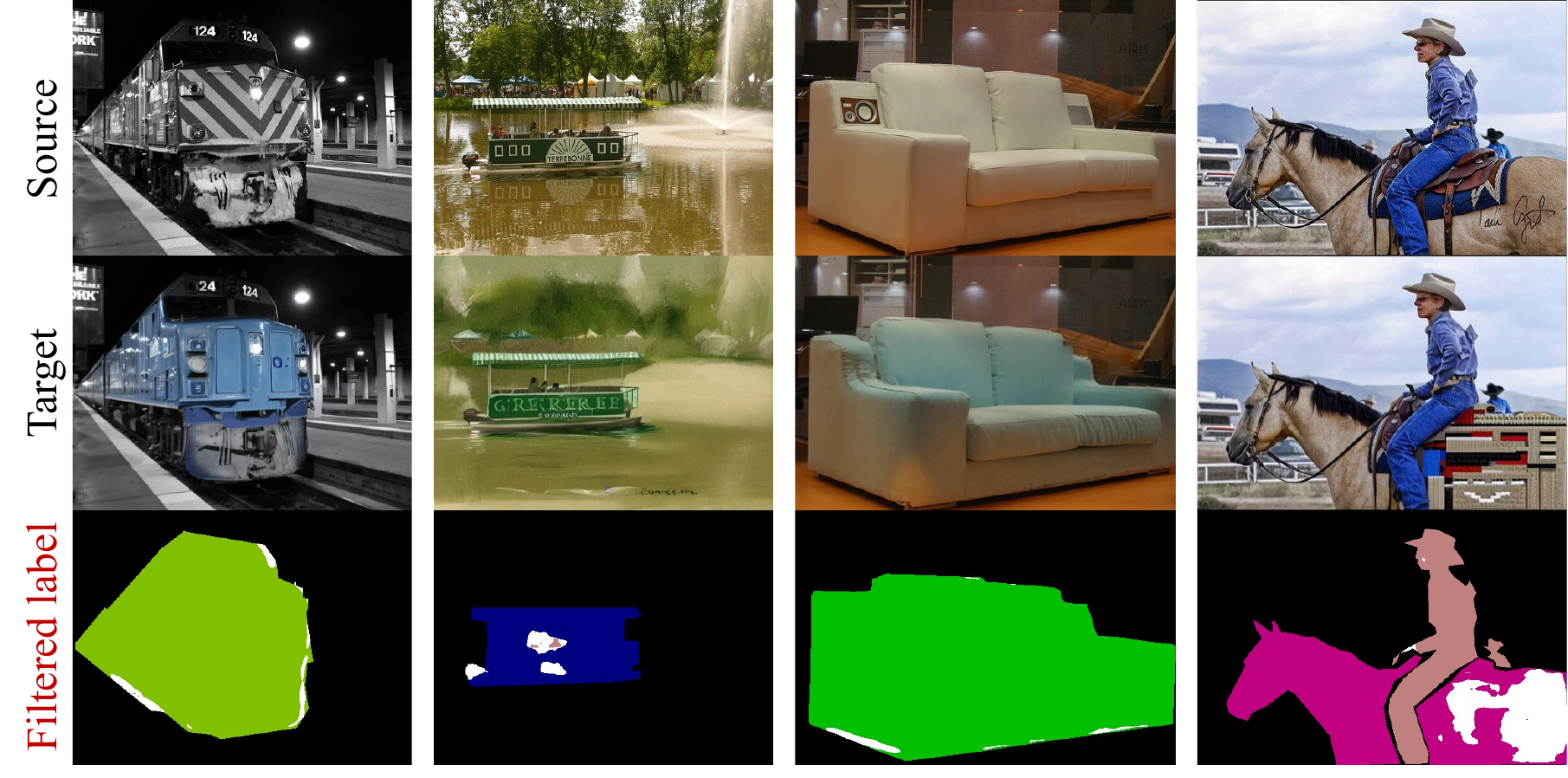}
    \caption{Visualization of synthetic images and the filtered regions (white masks in the label). The black masks are denoted as ``background'' category.}
    \vspace{-0.5em}
    \label{fig:filter_vis}
\end{figure}

\noindent \textbf{Noise Filtering Strategy.} 
In our noise filtering strategy, we utilize UperNet as a surrogate model to detect potential noisy regions and artifacts within the generated images. To demonstrate the effectiveness of this design, we visualize the filtered regions in Fig. \ref{fig:filter_vis}. These filtered regions exhibit several distinct patterns, such as object boundaries, synthesis failures, and nonsensical areas. This visualization highlights how our strategy effectively identifies and isolates problematic regions, ensuring higher quality and more reliable synthetic data for robustness evaluation.

\begin{figure}[t]
    \centering
    \includegraphics[width=\columnwidth]{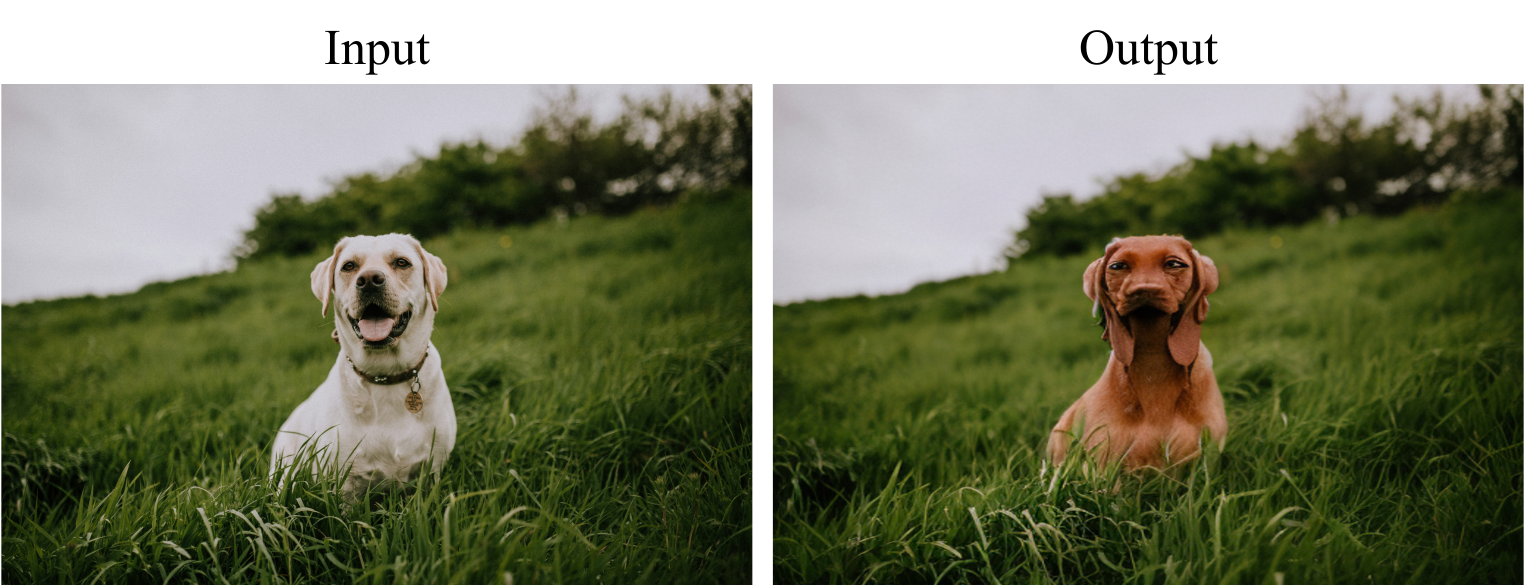}
    \caption{Failure cases of our proposed image editing pipeline.}
    \vspace{-1em}
    \label{fig:failure}
\end{figure}

\section{Limitations and Future Work}
\label{sec:limitations}
\noindent \textbf{Limitations.} First, current diffusion models occasionally struggle to preserve fine-grained details in the facial features of living subjects (\textit{e.g.}, animals and humans), which can compromise image realism, as shown in Fig. \ref{fig:failure}. To ensure the reliability of our benchmark, we currently address this by manually excluding samples containing front-facing biological subjects. Second, achieving fully disentangled attribute manipulation remains a challenge. Due to the inherent biases and spurious correlations within diffusion models, altering one attribute may inadvertently shift others; for instance, editing an object's material to 'wood' might unintentionally shift its color to 'brown'.

\noindent \textbf{Ethical Consideration.} While \textit{Gen4Seg} provides a scalable way to benchmark models, we acknowledge the ethical considerations associated with synthetic data. First, generative models may inherit and even amplify biases from their pre-training data, potentially leading to skewed appearance distributions or unintended attribute correlations. Second, the pipeline may encounter privacy and data copyright problems. To mitigate these risks, we emphasize that \textit{Gen4Seg} is designed as a diagnostic tool to complement, rather than replace, real-world datasets. For real-world applications, we advocate that insights derived from our synthetic benchmark should be cross-validated with real-world data and domain-specific protocols to ensure a comprehensive and responsible evaluation of model reliability.

\noindent \textbf{Future work.} We plan to address these generative limitations and extend our pipeline to cover additional variations, such as illumination changes, motion blur, and object rotation. Furthermore, we plan to apply our framework to a broader range of tasks, including instance segmentation, panoptic segmentation, and referring segmentation.

\section{Conclusion}
\label{sec:conclusion}
In this paper, we address the critical challenge of evaluating robustness against attribute variations in the semantic segmentation task. We introduce an automatic data generation pipeline, named \textit{Gen4Seg}, to synthesize diverse appearance and geometry variations from real images, in collaboration with diffusion, LLMs, and VLMs. To eliminate the need for re-annotating dense labels for generated samples, we propose a mask-guided energy function within the diffusion model to edit target objects while preserving the original image structures. To further enhance image quality, we introduce a two-stage automatic noise filtering strategy to reduce noise and artifacts in the generated images. This pipeline enables faithful attribute editing in a tuning-free manner, avoiding manual annotation of ground-truth segmentation masks. 
Using this pipeline, we construct two new benchmarks, Pascal-EA and COCO-EA. We evaluate the robustness of various semantic segmentation models, spanning different network architectures and recent open-vocabulary large models. We summarized several effective findings and validated that our pipeline can improve segmentation performances under both in-distribution and out-of-distribution data. Additionally, we demonstrate that the synthetic data generated by our diffusion models outperforms popular synthetic benchmarks in terms of image quality and reliability. We hope that our proposed pipeline and benchmarks will encourage further advancements in robust semantic segmentation.

\section{Acknowledgment}
This work was supported by the National Nature Science Foundation of China (Grant 62476029, 62225601, U23B2052), funded by the Fundamental Research Funds for the Beijing University of Posts and Telecommunications under Grant 2025TSQY08, the Beijing Natural Science Foundation Project No. L242025, the BUPT Excellent Ph.D. Students Foundation No. CX20242081, and sponsored by Beijing Nova Program and the Beijing Key Laboratory of Multimodal Data Intelligent Perception and Governance.

\bibliographystyle{IEEEtran}
\bibliography{reference}

\begin{IEEEbiography}
[{\includegraphics[width=1in,height=1.25in,clip,keepaspectratio]{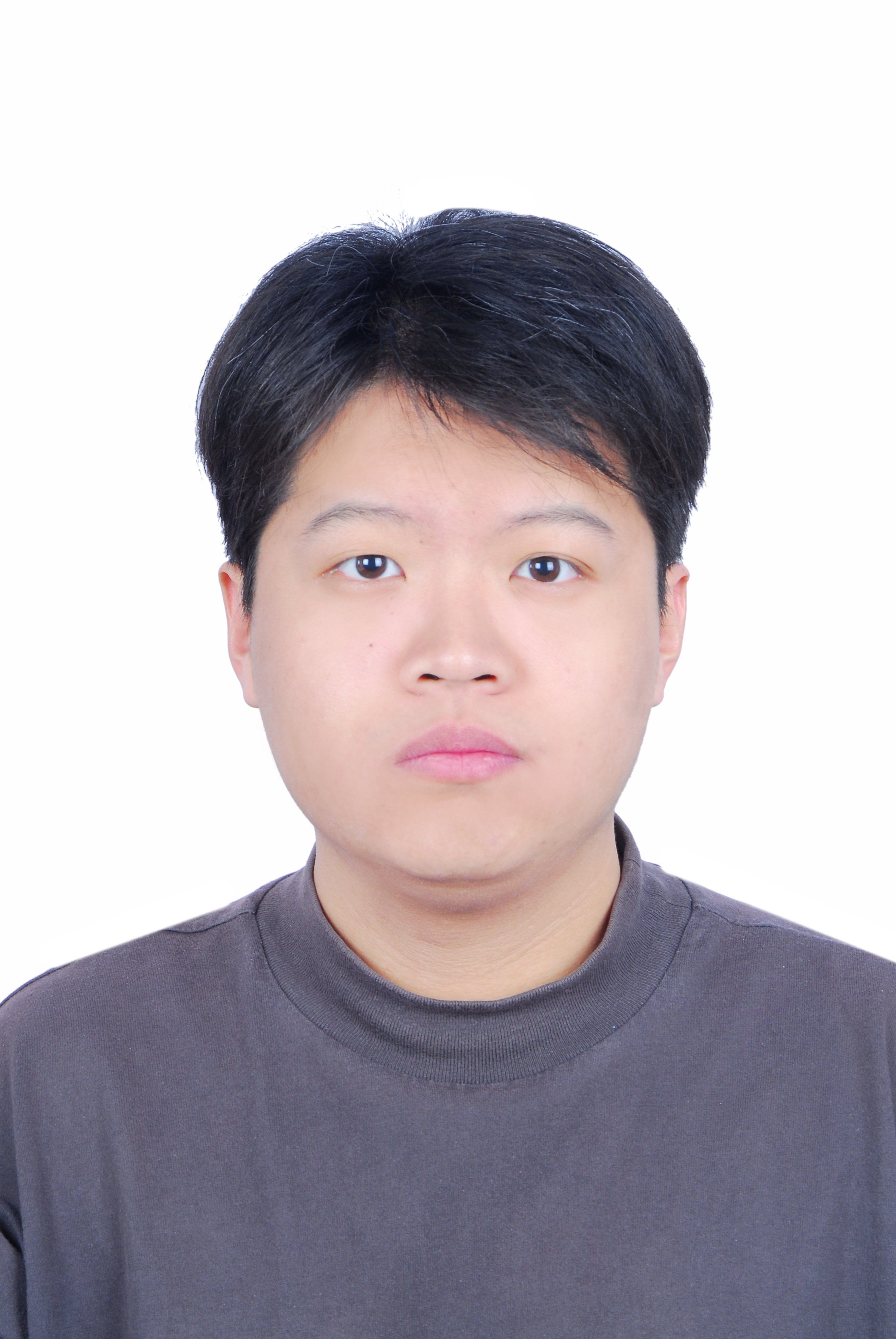}}]
{Zijin Yin} received the Bachelor's degree from Beijing University of Posts and Telecommunications, Beijing, China, in 2017. Currently, he is pursuing the Ph.D. degree at Beijing University of Posts and Telecommunications, Beijing, China. His research interests include diffusion models, semantic segmentation, and medical image analysis.
\end{IEEEbiography}

\begin{IEEEbiography}
[{\includegraphics[width=1in,height=1.25in,clip,keepaspectratio]{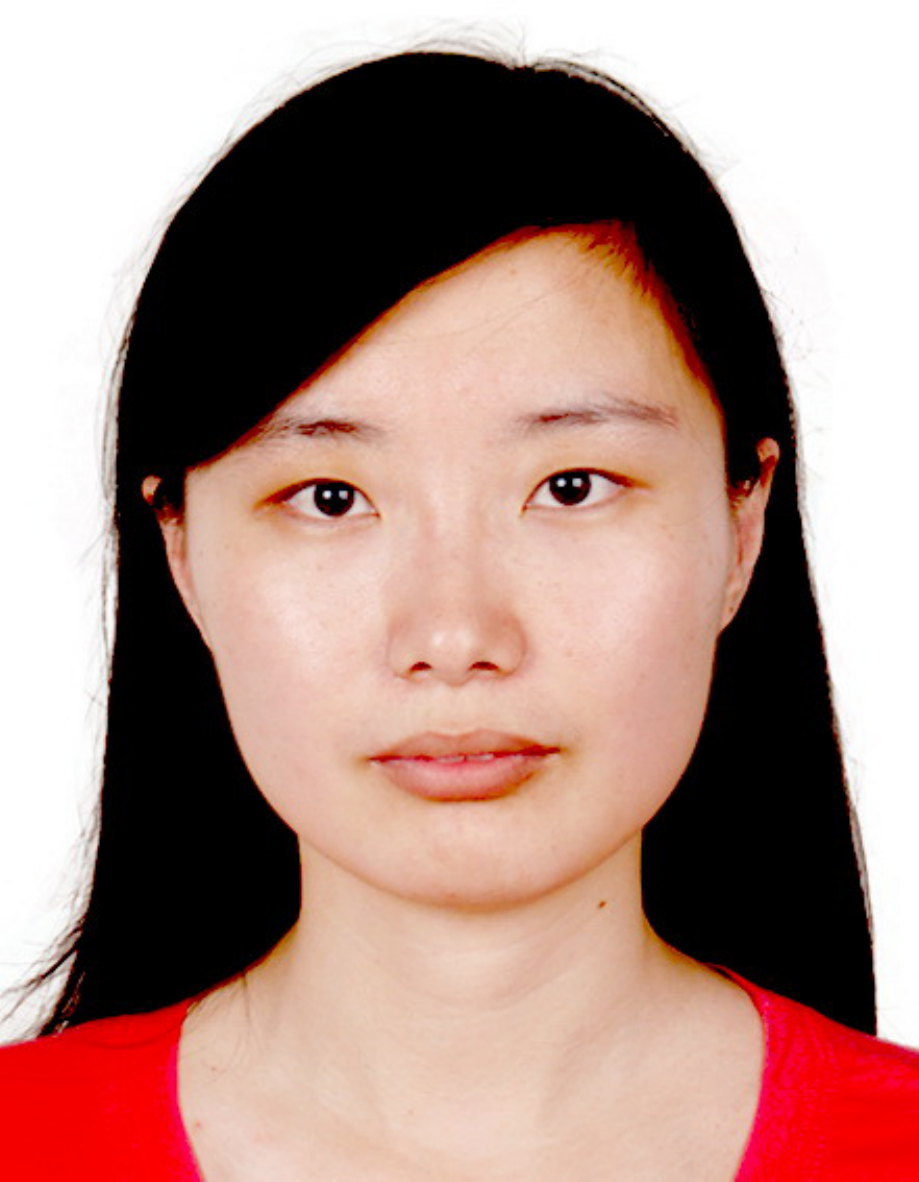}}]
{Bing Li} received a B.S. degree in Computer Science from Jinan University, Guangzhou, China, in 2009 and a Ph.D. degree from the Institute of Computing Technology, Chinese Academy of Sciences, Beijing, China, in 2016. Bing Li worked as a Postdoc Fellow at the University of Southern California, USA, in 2016.  She is currently a research scientist at King Abdullah University of Science and Technology. Her research interests include image/video processing, computer vision, and machine learning.
\end{IEEEbiography}

\begin{IEEEbiography}
[{\includegraphics[width=1in,height=1.25in,clip,keepaspectratio]{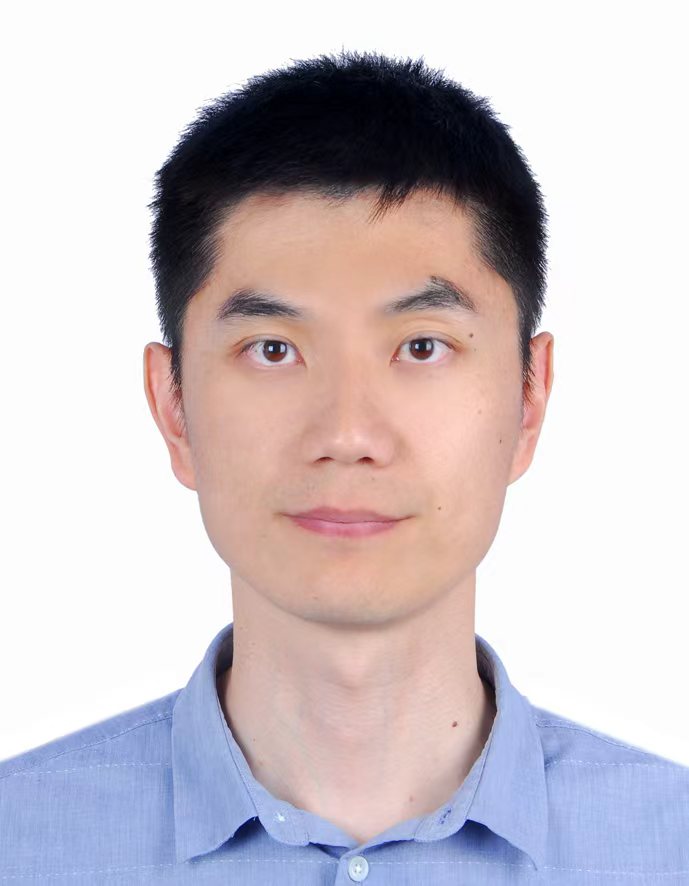}}]
{Kongming Liang} received the Bachelor's degree from China University of Mining \& Technology, Beijing, China, in 2012; and the Ph.D. degree from Institute of Computing Technology, Chinese Academy of Sciences, Beijing, China, in 2018. He was a joint Ph.D. Student of machine learning group in Carleton University from Sep 2016 to Oct 2017 and a postdoc researcher in the Department of Computer Science at Peking University from Jan 2019 to Dec 2020. Currently, he is an associate professor of Beijing University of Posts and Telecommunications. His research interests cover computer vision and machine learning, especially visual attribute learning, visual relationship detection, and medical image analysis.
\end{IEEEbiography}

\begin{IEEEbiography}
[{\includegraphics[width=1in,height=1.25in,clip,keepaspectratio]{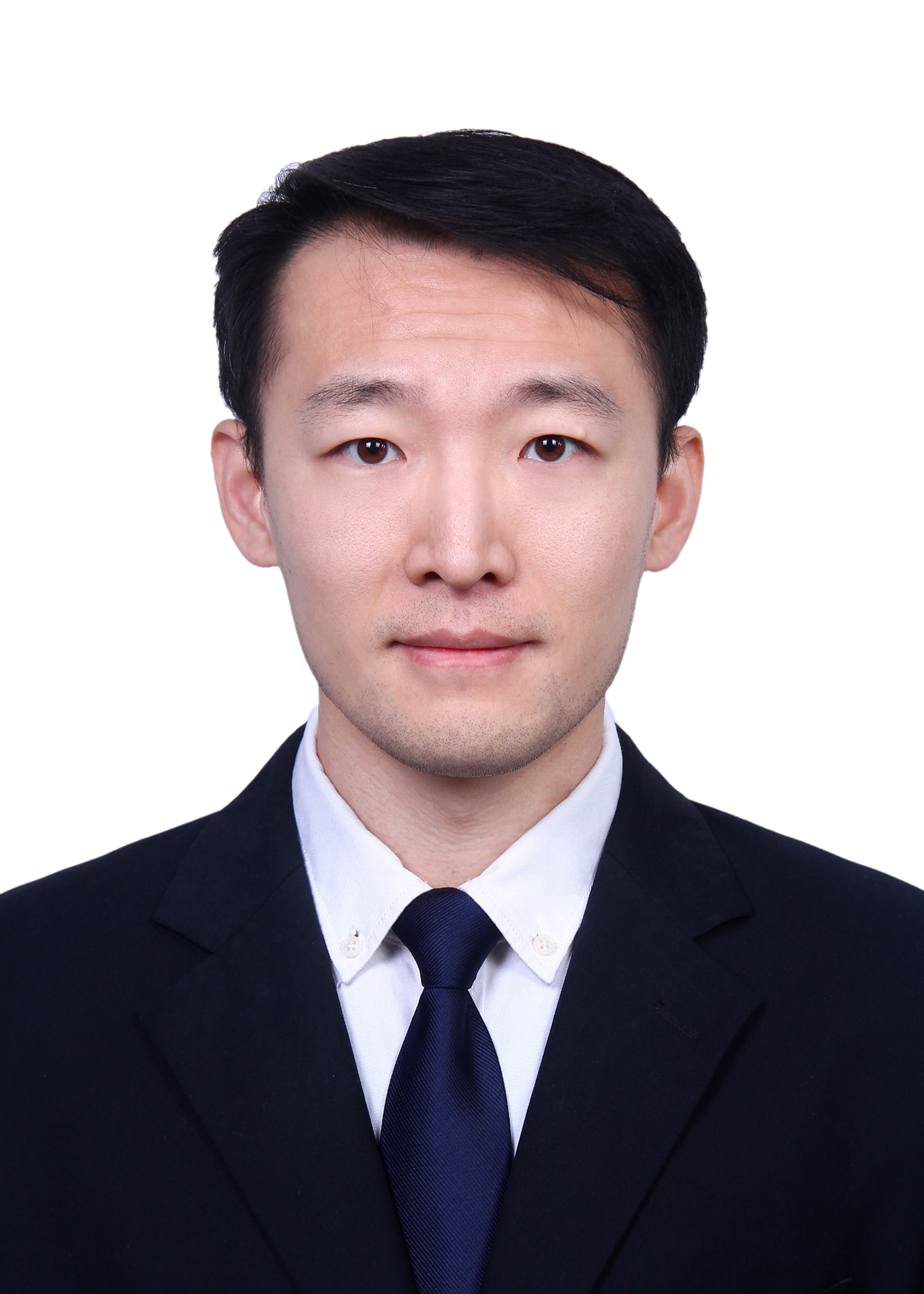}}]
{Hao Sun} received the PhD degree in signal and information processing from the University of Chinese Academy of Sciences, Beijing, in 2012. He is currently a senior researcher at China Telecom Artificial Intelligence Technology Co. Ltd. His research interests include computer vision, large multimodal models, and multimedia information retrieval.
\end{IEEEbiography}

\begin{IEEEbiography}
[{\includegraphics[width=1in,height=1.25in,clip,keepaspectratio]{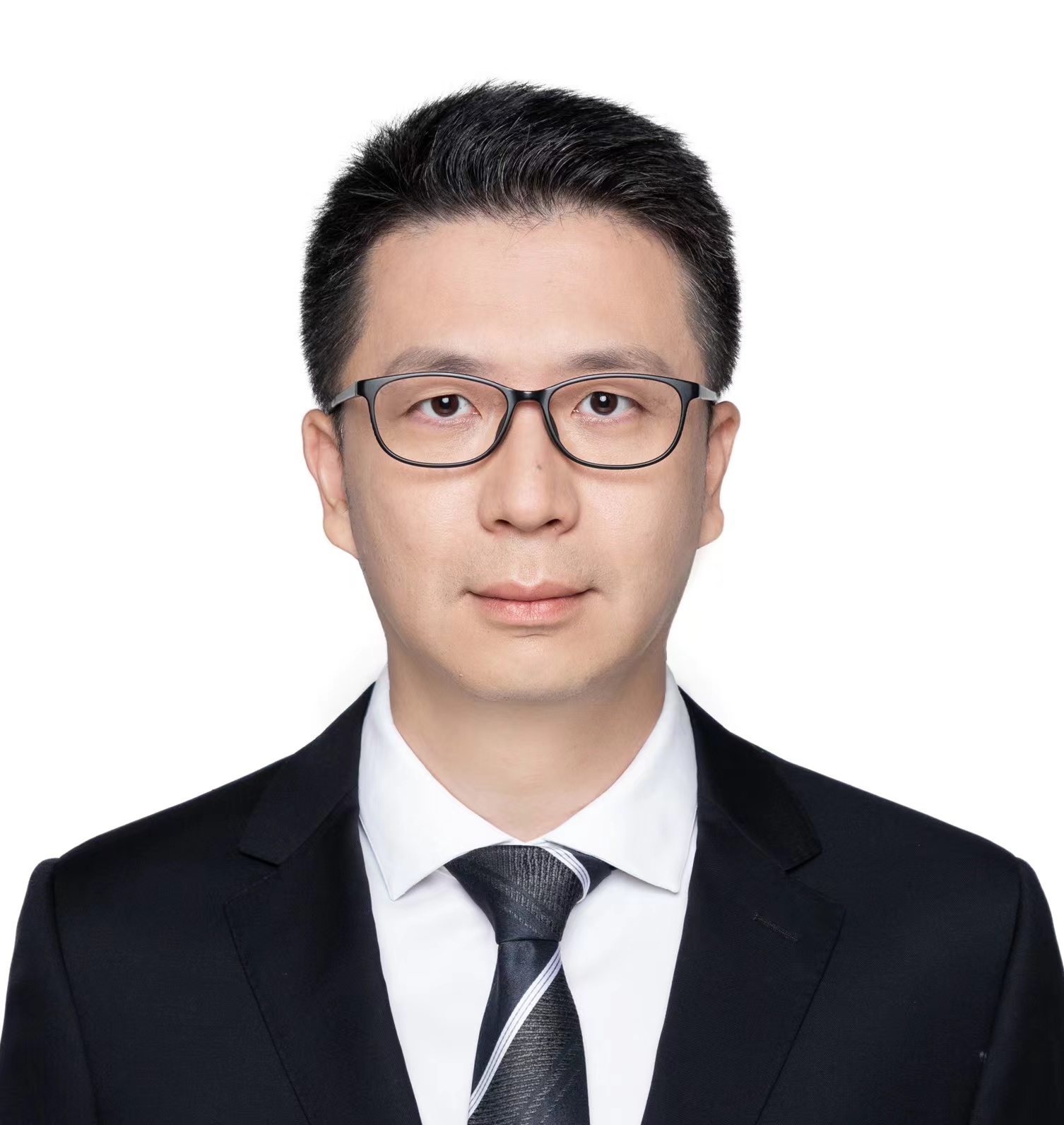}}]
{Zhongjiang He} is currently the Chairman and General Manager of China Telecom Artificial Intelligence Technology Co. Ltd. He received the B.E. degree in Optoelectronic Information Engineering from Zhejiang University, Hangzhou, China, in 2002, and the MBA from Tsinghua University, Beijing, China, in 2008. Since September 2022, he has been pursuing a Ph.D. degree in Electronic Information at Beijing University of Posts and Telecommunications, Beijing, China. His research interests include large multimodal models, multimedia information retrieval, retrieval-augmented generation, and image understanding.
\end{IEEEbiography}

\begin{IEEEbiography}
[{\includegraphics[width=1in,height=1.25in,clip,keepaspectratio]{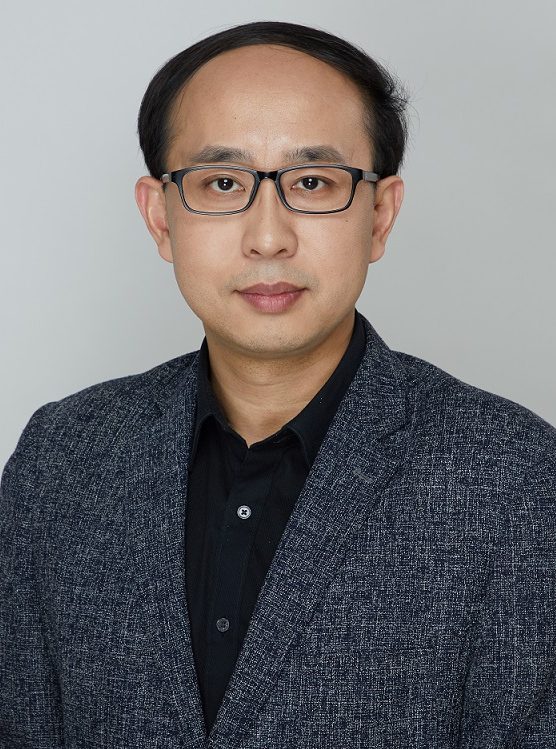}}] 
{Zhanyu Ma} is currently a Professor at Beijing University of Posts and Telecommunications, Beijing, China, since 2019. He received the Ph.D. degree in electrical engineering from KTH Royal Institute of Technology, Sweden, in 2011. From 2012 to 2013, he was a Postdoctoral Research Fellow with the School of Electrical Engineering, KTH. He has been an Associate Professor with the Beijing University of Posts and Telecommunications, Beijing, China, from 2014 to 2019. His research
interests include pattern recognition and machine learning fundamentals with a focus on applications in computer vision, multimedia signal processing. He is a Senior Member of IEEE.
\end{IEEEbiography}

\begin{IEEEbiography}
[{\includegraphics[width=1in,height=1.25in,clip,keepaspectratio]{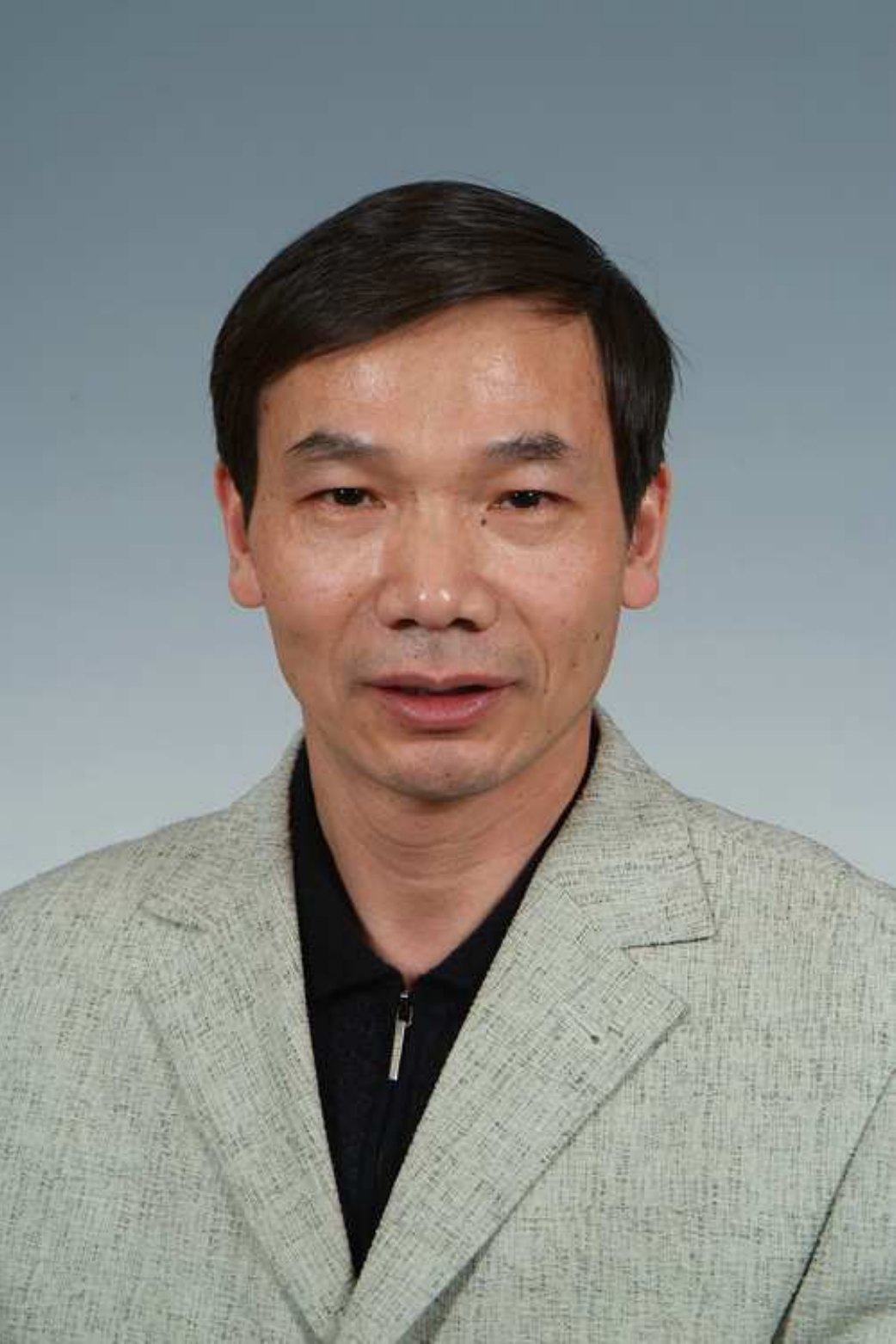}}]
{Jun Guo} received the B.E. and M.E. degrees from the Beijing University of Posts and Telecommunications (BUPT), China, in 1982 and 1985, respectively, and the Ph.D. degree from the Tohuku Gakuin University, Japan, in 1993. He is currently a professor and a vice president with BUPT. He has authored over 200 papers in journals and conferences, including Science, Nature Scientific Reports, the IEEE Transactions on PAMI, Pattern Recognition, AAAI, CVPR, ICCV, and SIGIR. His research interests include pattern recognition theory and application, information retrieval, content-based information security, and bioinformatics.
\end{IEEEbiography}
 
\vfill

\end{document}


\twocolumn[{%
\renewcommand\twocolumn[1][]{#1}%
\begin{center}
\large\vrule depth 0pt height 0.5pt width 2.3cm\hspace{0.1cm}Supplementary Material\hspace{0.1cm}\vrule depth 0pt height 0.5pt width 2.3cm \\
 \huge { Benchmarking Semantic Segmentation Models via Appearance and Geometry Attribute Editing } \vspace{-0.25cm} \\
 \IEEEcompsocdiamondline \\ \vspace{0.4cm}
\end{center}
}]

\begin{table*}[]
\renewcommand\arraystretch{1.00}
\centering
\caption{Summary of existing benchmarks for semantic segmentation robustness evaluation. Our benchmarks cover both global and local variations. We extend geometry attributes compared to our conference version.}
\label{tab:compare_benchmark}
\resizebox{0.9\textwidth}{!}{
\begin{tabular}{lccll}
\toprule
Dataset                             & Type     & Scene    & Global variation & Local variation\\
\midrule
Fog Cityscapes{\color{gray}\tiny [IJCV2018]} \cite{foggydriving}  & simulator   & cityscape   & fog & - \\
ACDC{\color{gray}\tiny [ICCV2021]} \cite{acdc}                    & real     & cityscape      & weather, daytime  &   -    \\
Dark Zurich{\color{gray}\tiny [TPAMI2020]} \cite{darkzurich}       & real    & cityscape      & daytime   & - \\
Multi-weather{\color{gray}\tiny [ICCV2021]} \cite{multiweather}    & GAN     & cityscape      & weather   &  -   \\
SHIFT{\color{gray}\tiny [CVPR2022]} \cite{shift}                  & simulator  & cityscape   & weather, daytime  & - \\
Stylized COCO{\color{gray}\tiny [ECCV2022]} \cite{trapped}        & AdaIN      & general     &  - & object texture \\
POC{\color{gray}\tiny [ECCV2024]} \cite{de2024placing}            & Diffusion  & general  & - & OOD objects \\
GenVal{\color{gray}\tiny [ECCV2024]} \cite{trapped}               & Diffusion     & cityscape    & weather, daytime & OOD objects \\
\midrule
Conference version{\color{gray}\tiny [CVPR2024]} \cite{yin2024benchmarking}  & Diffusion     & general    & weather, daytime, style  & object color, material\\   
Ours      & Diffusion & general & weather, daytime, style & object color, material, \textbf{size}, \textbf{position}\\
\bottomrule
\end{tabular}}
\end{table*}

\section{Difference with previous benchmarks}
As shown in Table \ref{tab:compare_benchmark}, unlike prior benchmarks, which only focus on global weather/daytime changes, add a few local out-of-distribution (OOD) objects, and are limited in cityscape scenarios, our work covers a wider spectrum of variations.
Methodologically, we contribute an end-to-end pipeline that uses multiple off-the-shelf models in concert for caption generation, attribute editing, and data filtering, enabling scalable and automatic construction of high-quality segmentation test sets.
Finally, our experiments provide actionable insights: local object appearance and geometry variations significantly affect segmentation robustness and should be explicitly considered in future evaluations and method design.

\section{Appearance Attribute Editing}

\noindent \textbf{Text editing.} 
For appearance attributes editing, we first instruct the Large Language Model (LLM) to edit text descriptions of source images. The detailed prompts and exemplar inputs and outputs are shown in Tab. \ref{tab:prompts}.

\begin{table*}[t]
\centering
\caption{Illustrations of text prompts used to instruct LLaMA3 \cite{llama3} and edited text examples. }
\label{tab:prompts}
\renewcommand\arraystretch{1.5}
\resizebox{\textwidth}{!}{
\begin{tabular}{m{2cm}|p{12cm}|p{7cm}}
\toprule
\textbf{Attribute} & \textbf{Instruction prompt} & \textbf{Examples} \\
\hline
Color         &  \textit{I want to change the color of the object in the source image. Please generate all possible target text prompts given the source text prompt describing the source image. For example, the source is ``a cat'', you can generate ``a blue cat''.}  & Input: \textit{a white and red train.} Output: \textit{a \textcolor{red}{white and blue} train.}  \\
\hline
Material      &  \textit{I want to change the material of the object in the source image. Please generate all possible target text prompts given the source text prompt describing the source image. For example, source is ``a cat'', you can generate ``a wooden cat sculpture''.}  &  Input: \textit{an airplane on the ground.} Output: \textit{a \textcolor{red}{wooden airplane model} on the ground.} \\
\hline
Style:        & \textit{I want to change the image style of source images without perturbing the content. Please generate all possible target text prompts given the source text prompt describing the source image. For example, source is 'a cat', you can generate 'a watercolor cat'.} & Input: \textit{A brown leather couch is sitting outside on a sidewalk.} Output: \textit{A \textcolor{red}{digital painting of} a brown leather couch is sitting outside on a sidewalk.}    \\
\hline
Weather:      & \textit{I want to change the weather or season condition of the source image. Please generate all possible target text prompts given the source text prompt that describes the source image, only changing the weather conditions, or adding a description of the weather if not already present.} & Input: \textit{Two sheep lying in the grass.} Output: \textit{Two sheep laying in the grass \textcolor{red}{under a heavy downpour}.} \\
\bottomrule
\end{tabular}}
\end{table*}

\noindent \textbf{Mask-Guidance Diffusion.} The algorithm procedure of our proposed Mask-Guidance Diffusion is shown in Alg. \ref{alg:appear_edit}. 

\section{Datasets Setup}
\noindent \textbf{Implementation Details.} We apply Stable-Diffusion-2-1-base as a base model for appearance attribute editing. We use 50 time-steps for both DDIM inversion and sampling. We set $\theta_f$ to $0.8$ and $\theta_a$ to $0.5$ for features and self-attention injections. For our mask-guidance energy function loss, we set the loss threshold $\gamma$ to $0.2$ and the loss scale factor $\eta$ to $1$. In object geometry editing, we use the Stable-Diffusion-Inpainting model \cite{ldm}. We employ the naive UperNet \cite{upernet} as a surrogate model for pixel-level noise filtering. All experiments are conducted on four NVIDIA RTX 4090Ti. 

\noindent \textbf{Dataset Statistics}
It is noteworthy that our primary objective is to provide a comprehensive benchmark for evaluating segmentation models’ capabilities under specific attribute changes. The final dataset preparation involved a human-in-the-loop process for both intermediate and final cleaning, where we manually filtered the highest-quality images. In Pascal-EA, we collected 805 images across the color, material, weather, and style subsets, and 848 images for the size and position subsets. In COCO-EA, there are 2,879 images in the color, material, weather, and style subsets, and 3,034 images in the size and position subsets.

\section{More benchmark results} 
\begin{figure*}[t]
    \centering
    \includegraphics[width=\linewidth]{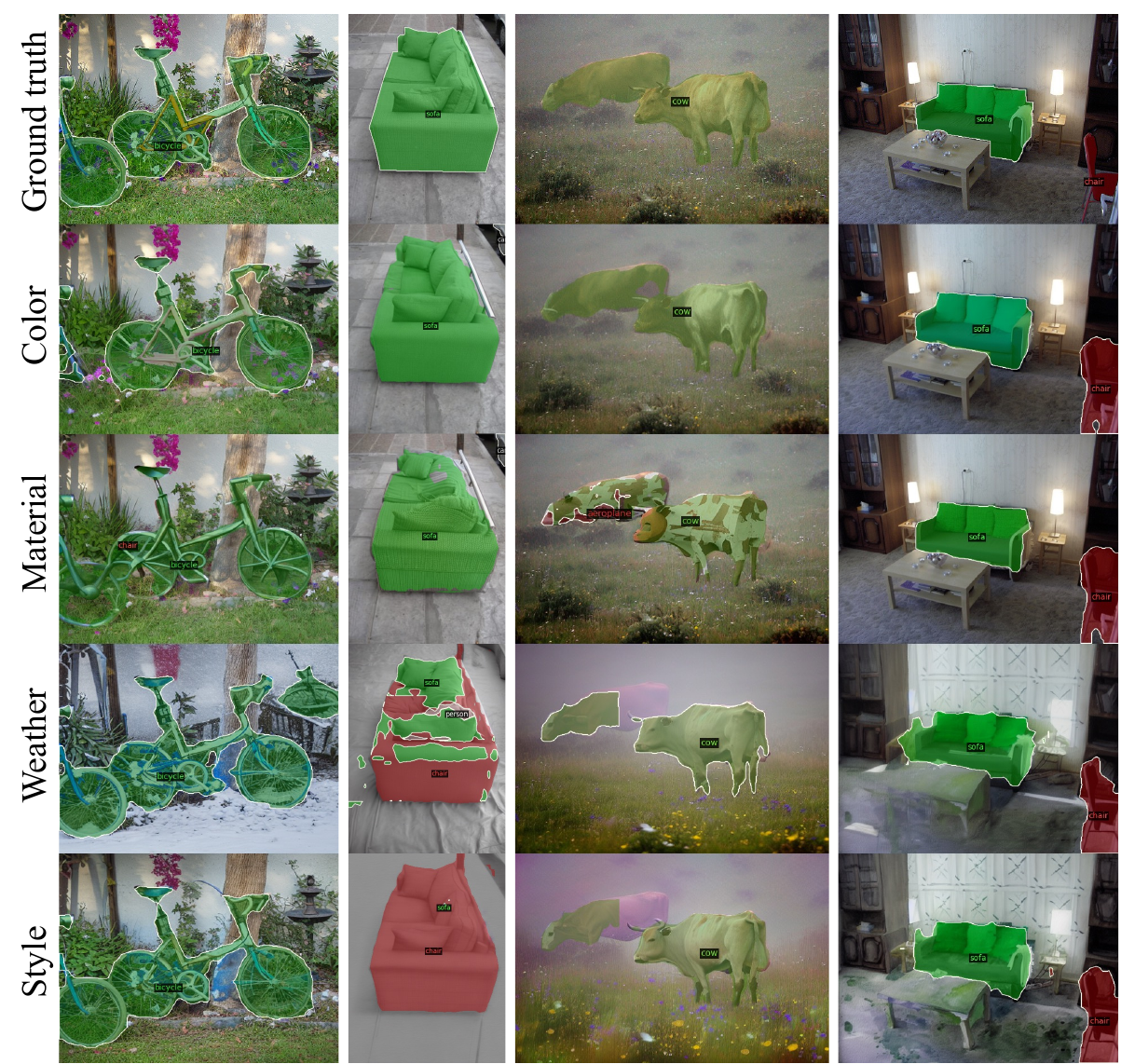}
    \caption{Visualization results under appearance variations on our proposed Pascal-EA benchmarks.}
    \label{fig:seg_vis_app_pas}
\end{figure*}

\begin{figure*}[t]
    \centering
    \includegraphics[width=\linewidth]{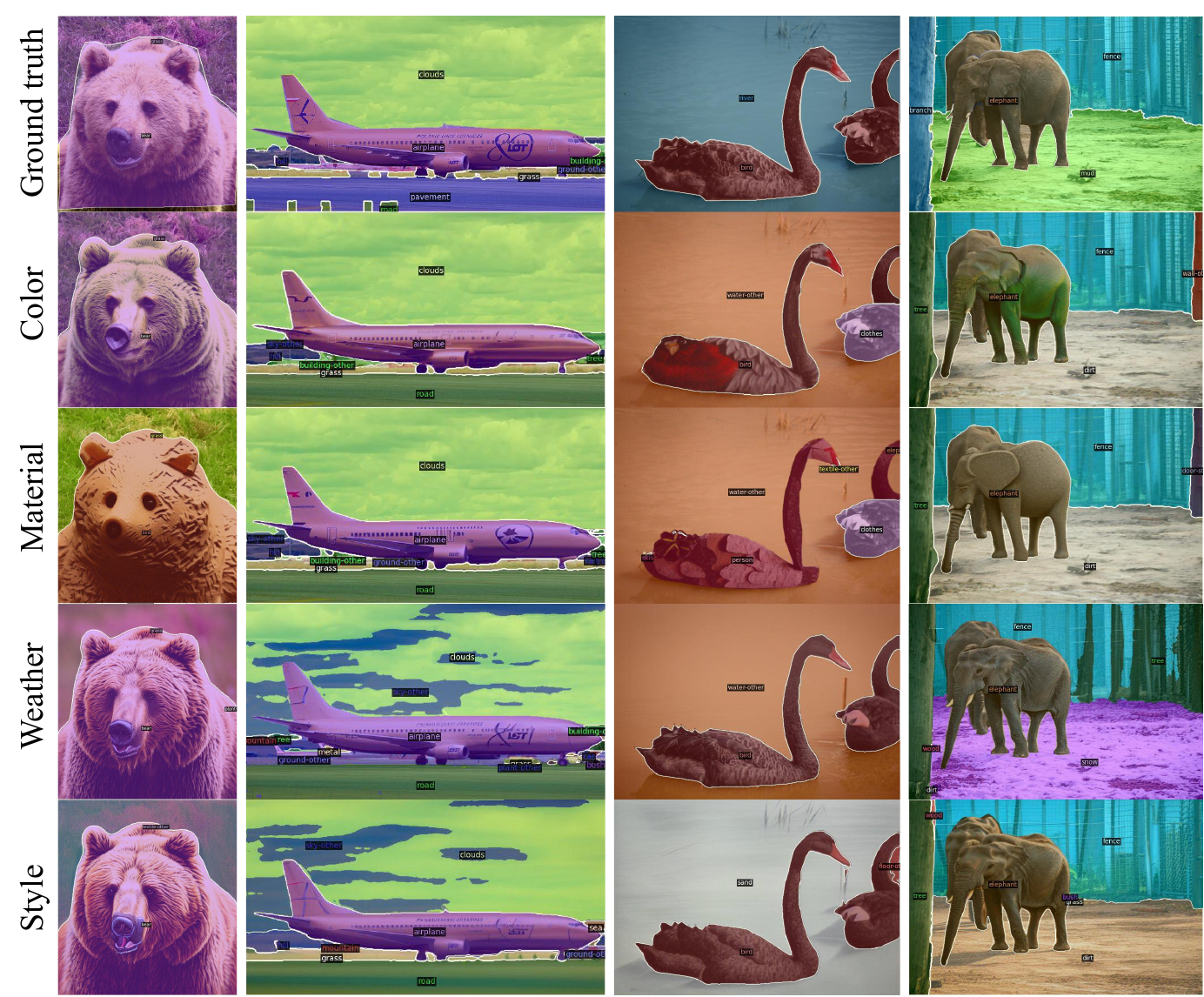}
    \caption{Visualization results under appearance variations on our proposed Pascal-EA benchmarks.}
    \label{fig:seg_vis_app_coco}
\end{figure*}

\begin{figure*}
    \centering
    \includegraphics[width=\linewidth]{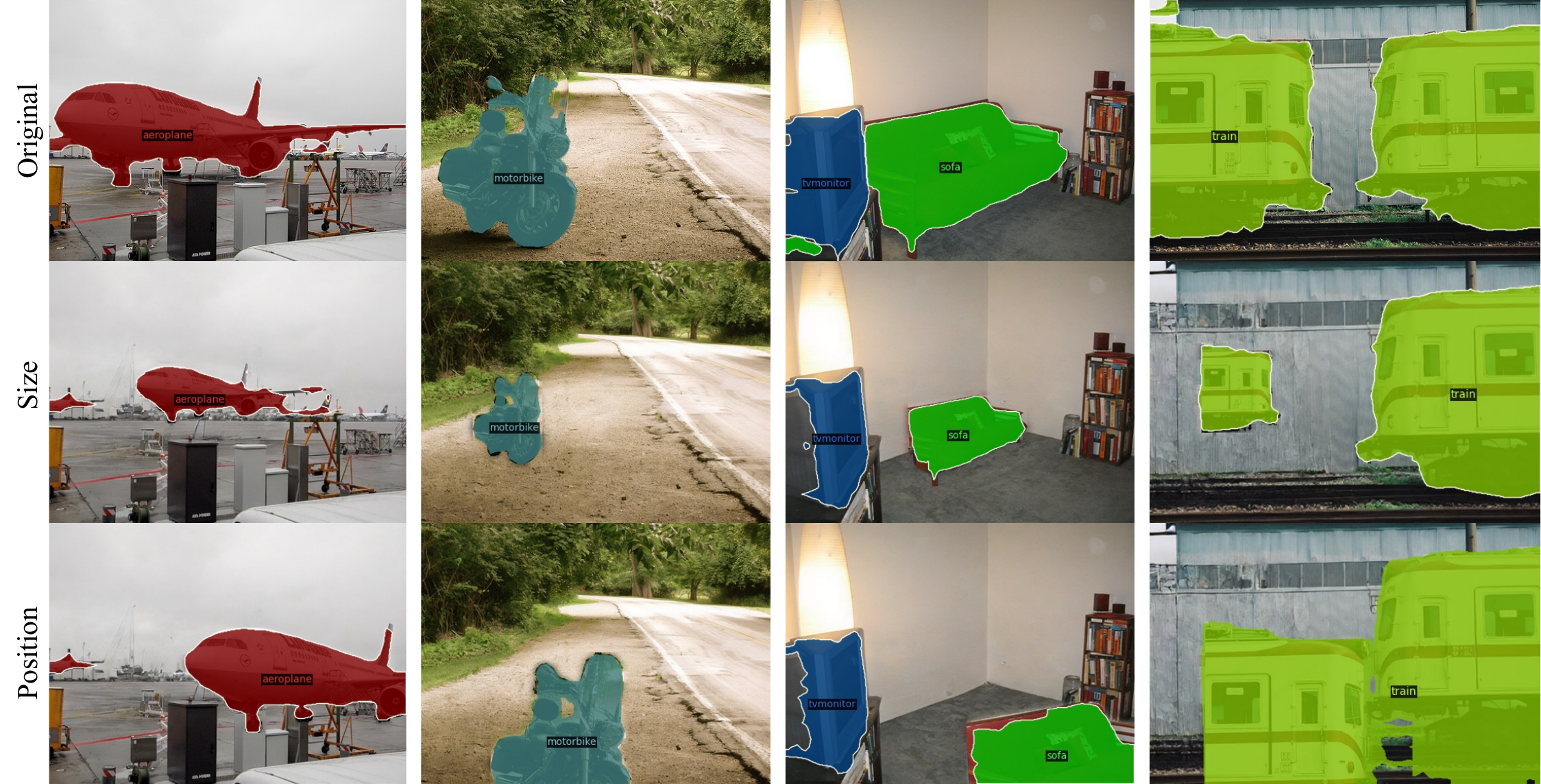}
    \caption{Visualization results under object geometry (size and position) variations.}
    \label{fig:seg_vis_geo}
\end{figure*}

Qualitative results under attribute changes are shown in Fig. \ref{fig:seg_vis_app_pas}, Fig. \ref{fig:seg_vis_app_coco}, and Fig. \ref{fig:seg_vis_geo}.

\textbf{Finding-8:} \textit{Local attribute variations can pose considerable challenges to segmentation performance comparable to those caused by global changes.}

The Tab. 1 from the main paper illustrates that the average mIoU values under local color alteration reach 80.47\% and 41.92\%, while the values under global weather and style changes are 79.64\%/76.15\% and 37.47\%/35.59\% on Pascal-EA and COCO-EA, respectively. These results show that segmentation performance under color variation is slightly higher than under style and weather variations.
Similarly, the Tab. 2 from the main paper reveals that average mIoU values under local size-0.2 and position-0.2 variations are 71.25\%/70.92\% and 21.82\%/22.43\% on Pascal-EA and COCO-EA, respectively. These results indicate that local changes can pose considerable challenges to segmentation models, comparable to global variations. Given that local modifications affect only small portions of the image, their significant impact underscores the importance of testing against these variations.

\textbf{Finding-9:} \textit{All methods have performance decay as the geometry editing becomes heavier.}

The Tab. 2 from the main paper shows that PSPNet drops from 64.97\% to 62.83\%, and from 64.98\% to 63.67\% as editing strengths of size and position increase from 0.2 to 0.4 on Pascal-EA, respectively. And average mIoU values in the last row also support this trend. This reveals that model performance decays as the editing becomes heavier. The observation also suggests that, unlike adversarial attack methods, our pipeline does not cause models to collapse. Instead, it provides a robust means for continuous stress-testing of segmentation models, offering a balanced evaluation across geometric alterations.

\textbf{Finding-10:} \textit{When two types of variation are combined, the overall performance deteriorates more compared to when only one variation is present.}

\begin{figure}[t]
    \centering
    \includegraphics[width=\linewidth]{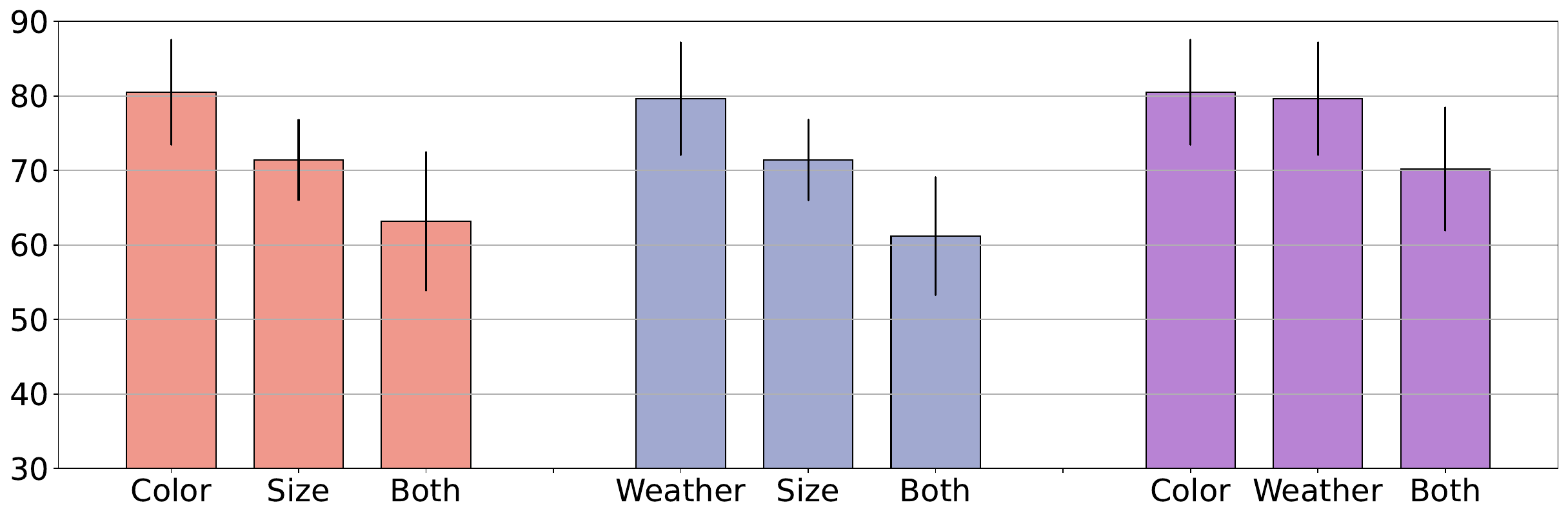}
    \caption{Average mIoU ($\uparrow$) of segmentation methods under two different attribute variations and their combinations.}
    \vspace{-0.5em}
    \label{fig:multiedits}
\end{figure}

Since our generative benchmark is extensible, we can also stress-test models when exposed to samples with a combination of two distinct attribute variations. 
The results are presented in Fig. \ref{fig:multiedits}. The average mIoU is around 80\% and 70\% under object color and size variations, respectively, but drops to approximately 60\% when both variations are combined. A similar trend is observed with the combination of global and local variations. This observation indicates that addressing multiple attribute variations simultaneously is more challenging than dealing with them in isolation, suggesting that models struggle more with compounded variations.

\begin{algorithm}[]
\caption{Image Editing with Mask-Guidance Diffusion}
\begin{algorithmic}[1]
\setstretch{1.0}
\State \textbf{Inputs:} Source real image $I$, source and target prompt $P$ and $P^*$, object mask $M$, injection thresholds $\theta_f$ and $\theta_a$, loss threshold $\gamma$ and loss scale factor $\eta$.
\State \textbf{Outputs:} Source latent map $z_0$ and target latent map $z_0^*$ corresponding to $P$ and $P^*$.
\State $z_T^G \gets DDIMinversion(I)$ 
\State $z_T \gets z_T^G$    \Comment{Source branch}
\State $z_T^* \gets z_T^G$  \Comment{Target branch}
\For{$t \gets T, T-1 \ldots 1$}
    \State $\mathcal{L} \gets \infty$
    \While{$\mathcal{L} > \gamma$}
        \State $\epsilon_{s}^*, f_t^*, \widetilde{A}_t^*, \hat{A}_t^* \gets \epsilon_\theta(z_t^*, P^*, t)$
        \State $\mathcal{L} \gets EnergyFunction(\hat{A}_t^*, M)$ 
        \State $z^*_t \gets z^*_t - \eta \nabla_{z^*_t} \mathcal{L}$  \Comment{Gradient guidance}
    \EndWhile
    \State $\epsilon_{s}, f_t, \widetilde{A}_t, \hat{A}_t \gets \epsilon_\theta(z_t, P, t)$
    \State $\epsilon_{s}^*, f_t^*, \widetilde{A}_t^*, \hat{A}_t^* \gets \epsilon_\theta(z_t^*, P^*, t)$
    \State \textbf{if}~{$t < \theta_f$}~~\textbf{then}~{$f_t^* \gets f_t$} \Comment{Feature injection}
    \State \textbf{if}~{$t < \theta_a$}~~\textbf{then}~{$\widetilde{A}_t^* \gets \widetilde{A}_t$} \Comment{Self-attention injection}
    \State $\epsilon_{s}^* \gets \epsilon_\theta(z_t^*, P^*, t, \{f_t^*, \widetilde{A}_t^*, \hat{A}_t^*\})$
    \State $z_{t-1}^* \gets DDIMsample(z_t^*, \epsilon_{s}^*)$ 
    \State $z_{t-1} \gets DDIMsample(z_t, \epsilon_{s})$
    \State $z^*_{t-1} \leftarrow M \odot z^*_{t-1} + (1 - M) \odot z_{t-1}$
\EndFor
\State \textbf{Return} $z_0, z_0^*$
\end{algorithmic}
\label{alg:appear_edit}
\end{algorithm}

\section{More Discussion and Analysis}

\noindent \textbf{Comparison with other Benchmarks.} 
Furthermore, since the previous Stylized COCO \cite{trapped} generates local style variations on objects to study robustness in segmentation, we also compare our method with it. Following its setting, we generate images where objects and backgrounds have random art styles, respectively. From the results in Table \ref{table:benchmark_compare2}, our method significantly surpasses it in image reality and structural preservation. 
Besides, we notice that previous stylized datasets have considerably greater FID scores, which implies that they could bring more out-of-distribution interference, and hence impede the reliability of their evaluation.

\begin{table*}[t]
\renewcommand\arraystretch{1.05}
\centering
\caption{Benchmarking the robustness of several methods under greater object size attribute changes on our Pascal-EA and COCO-EA datasets. The results are reported using mIoU ($\uparrow$). } 
\label{tab:benchmark_geo_size}
\resizebox{\textwidth}{!}{
\begin{tabular}{lc|cccc|cccc}
\toprule[1.28pt]
\multirow{2}{*}{Method} & \multirow{2}{*}{Backbone} & \multicolumn{4}{c|}{Pascal-EA}  & \multicolumn{4}{c}{COCO-EA}  \\ 
\cmidrule(lr){3-6} \cmidrule(lr){7-10}
&    & Original & Size 0.6 & Size 0.8 & $mR$ & Original & Size 0.6 & Size 0.8 & $mR$ \\ 
\midrule
DeepLabV3+{\color{gray}\tiny [CVPR2018]} \cite{deeplabv3plus} & ResNet50 & 66.69 & 30.21 & 10.25 & 0.30 & 22.07 & 17.12 & 10.45 & 0.62\\
Segmenter{\color{gray}\tiny [ICCV2021]} \cite{segmenter} & ViT-B/16 & 70.87 & 35.58 & 12.79 & 0.34 & 23.65 & 15.29 & 11.79 & 0.57 \\
Mask2Former{\color{gray}\tiny [CVPR2022]} \cite{mask2former}  &  Swin-B & 77.11 & 36.70 & 18.22 & 0.36 & 21.81 & 13.99 & 10.33 & 0.56 \\
OVSeg{\color{gray}\tiny [CVPR2023]} \cite{ovseg} &  Swin-B  & 79.46 & 40.29 & 20.85 & 0.38 & 23.42 & 15.79 & 11.74 & 0.59\\
SEEM{\color{gray}\tiny [NeurIPS2023]} \cite{seem}    &  Focal-T  & 76.24 & 37.29 & 20.53 & 0.38 & 27.38 & 16.38 & 13.22 & 0.54
\\
\bottomrule[1.28pt]
\end{tabular}}
\end{table*}

\begin{table}[]
\centering
\caption{Comparison of different surrogate models used for noise filtering.} 
\label{tab:surrogate}
\resizebox{\columnwidth}{!}{
\begin{tabular}{lcccc}
\toprule[1.28pt]
\multirow{3}{*}{Method}  & \multicolumn{2}{c}{\textbf{PASCAL VOC}} & \multicolumn{2}{c}{\textbf{COCO Stuff 164k}} \\ 
\cmidrule(lr){2-3}  \cmidrule(lr){4-5} 
& Filtered/Total &  CLIP Accuracy  & Filtered/Total &  CLIP Accuracy \\
\midrule
UperNet \cite{upernet}    & 14/200  & 0.879 &  12/200  & 0.882 \\
OCRNet \cite{ocr} & 14/200  & 0.871 & 12/200  & 0.890 \\
Segmenter \cite{segmenter} & 19/200  & 0.884 & 15/200  & 0.896 \\
Mask2Former \cite{mask2former}   &  23/200 & \textbf{0.903} & 18/200  & \textbf{0.915} \\
\bottomrule[1.28pt]
\end{tabular}}
\end{table}

\begin{figure}
    \centering
    \includegraphics[width=\linewidth]{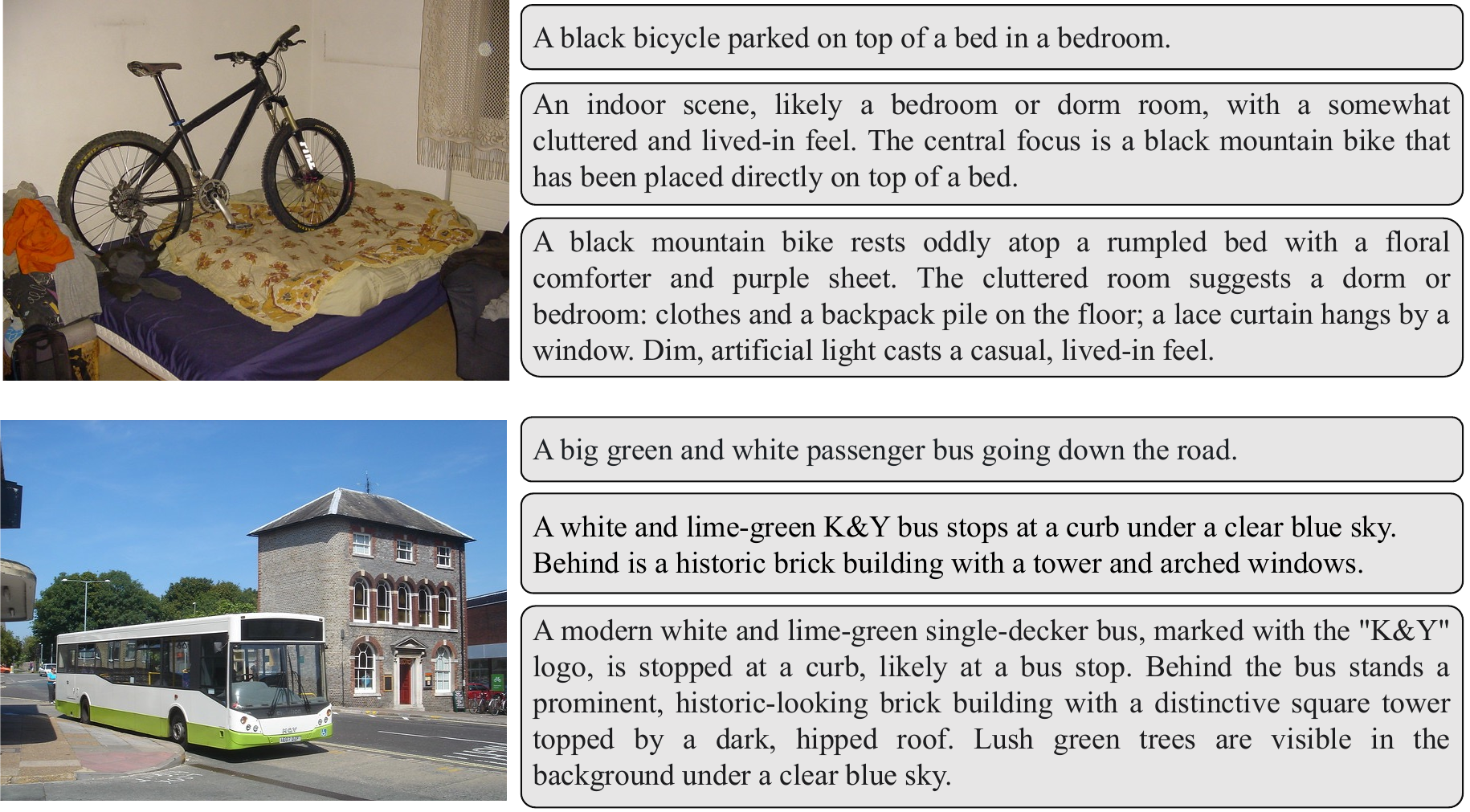}
    \caption{Two examples of captions with varying levels of detail. From top to bottom are short, medium, and long captions, respectively.}
    \label{fig:caption_vis}
\end{figure}

\noindent \textbf{Ablation Study of captions with different lengths.}
We analyze the sensitivity of our framework to caption length, since text descriptions are used to guide the diffusion model for appearance and geometry editing. We select three caption variants with different lengths and levels of specificity, including short captions (adopted in the main manuscript), as well as medium and long captions with richer contextual and semantic details. All captions are generated by Qwen2-VL \cite{Qwen2VL}, and examples are shown in Fig. \ref{fig:caption_vis}. 
The quantitative results are reported in Tab. \ref{tab:ablation_caption}. We observe that short captions consistently achieve the best performance under both appearance and geometry editing scenarios. This is because concise captions focus on the core attributes of the target object, providing clearer and less ambiguous guidance for downstream editing, whereas longer captions may introduce redundant contextual information. 
We emphasize that the robustness to caption length is largely governed by the diffusion model. In our implementation, we adopt Stable Diffusion 2 as the base model, which is known to respond more favorably to short, object-centric prompts than to long, compositional descriptions. Thus, simple and concise captions yield more faithful and stable editing results.

\begin{table}[]
\centering
\caption{Comparison of our pipeline with Stylized COCO \cite{trapped} in local object and background changes. Our generated images have superior image reality and fidelity.} 
\label{table:benchmark_compare2}
\resizebox{\columnwidth}{!}{
\begin{tabular}{lcccc}
\toprule[1pt]
  & Model   & \makecell{CLIP \\Acc ($\uparrow$)} & \makecell{DINO \\Dist ($\downarrow$)}  \\
\midrule
Stylized-Object{\color{gray}\tiny [ECCV2022]} \cite{trapped}  & AdaIN &   0.574  &  0.075   \\
\rowcolor{gray!40}
Ours      & Diffusion &   \textbf{0.906}     &   \textbf{0.002}    \\
\midrule
Stylized-Background{\color{gray}\tiny [ECCV2022]} \cite{trapped} & AdaIN &  0.678  &   0.084   \\
\rowcolor{gray!40}
Ours      & Diffusion &  \textbf{0.975}    &  \textbf{0.004}    \\
\bottomrule[1pt]
\end{tabular}}
\end{table}

\noindent \textbf{Ablation Study of different surrogate models.} 
We selected four representative models: 1) UperNet \cite{upernet}, and OCRNet \cite{ocr}, which use a naive CNN backbone, and 2) Segmenter \cite{segmenter} and Mask2Former \cite{mask2former}, which use a more powerful vision transformer backbone.
We compared their effectiveness in noise detection by discarding samples where localized regions occupy more than 10\% of the total area, then computing the CLIP Accuracy between the filtered images and the originals. Higher CLIP scores indicate more effective filtering. Tab. \ref{tab:surrogate} shows that the stronger segmentation model filters more samples, but the improvement in CLIP Accuracy is slight.
These results validate that our system is robust to different surrogate models.

\noindent \textbf{Benchmark under greater geometry attribute changes.} 
To further probe model differences, we extended the scale of object size reduction beyond 0.4 to include more extreme reductions (0.6 and 0.8). The results in Tab. \ref{tab:benchmark_geo_size} show that, even under these stronger perturbations, the different methods produce comparable $mR$ values. This finding is consistent with our prior observations and previous research \cite{li2023imagenet}, that higher accuracy from more powerful architectures or larger training sets does not guarantee greater robustness.



\begin{table}[t]
\centering
\caption{Quantitative results of edit performance under different prompt lengths. The best results are in bold.}
\label{tab:ablation_caption}
\resizebox{\linewidth}{!}{
\begin{tabular}{lc ccc cc}
\toprule[1pt]
\multirow{2}{*}{Caption}  & \multirow{2}{*}{\makecell{Length\\(\#tokens)}}
& \multicolumn{3}{c}{Appearance} 
& \multicolumn{2}{c}{Geometry} \\
\cmidrule(lr){3-5} \cmidrule(lr){6-7}
& & \makecell{DINO \\Dist($\downarrow$)} & \makecell{Masked \\LPIPS($\downarrow$)} & \makecell{CLIP \\Score($\uparrow$)}
& FID($\downarrow$) & HPSv2($\uparrow$) \\
\midrule
Short   & $\approx$12 & \textbf{0.109} & \textbf{0.192} & \textbf{0.512} & \textbf{16.84} & \textbf{22.79} \\
Medium  & $\approx$30 & 0.118 & 0.207 & 0.395 & 17.92 & 19.38 \\
Long    & $\approx$60 & 0.242 & 0.346 & 0.171 & 35.38 & 18.26 \\
\bottomrule[1pt]
\end{tabular}}
\end{table}

\newpage

\newpage
\bibliographystyle{IEEEtran}
\bibliography{reference}

\newpage

\newpage

